\documentclass[10pt,twocolumn,letterpaper]{article}

\usepackage{iccv}              

\usepackage{url}
\usepackage{graphicx}
\usepackage{booktabs}
\usepackage{wrapfig}   
\usepackage{multirow}
\usepackage{diagbox}
\usepackage{adjustbox}
\usepackage{makecell}
\usepackage{colortbl}

\newcommand{\tabincell}[2]{\begin{tabular}{@{}#1@{}}#2\end{tabular}}
\newcommand{\rot}[2][50]{%
  \rotatebox{#1}{#2}%
}
\definecolor{iccvblue}{rgb}{0.21,0.49,0.74}
\usepackage[pagebackref,breaklinks,colorlinks,allcolors=iccvblue]{hyperref}

\def\paperID{****} 
\def\confName{ICCV}
\def\confYear{2025}

\title{Progressive Scaling Visual Object Tracking}

\author{Jack Hong\textsuperscript{\rm 1} ~
        Shilin Yan\textsuperscript{\rm 1} ~ Zehao Xiao\textsuperscript{\rm 2} ~ Jiayin Cai\textsuperscript{\rm 1} ~
        Xiaolong Jiang\textsuperscript{\rm 1} \\
        ~ Yao Hu\textsuperscript{\rm 1} ~ Henghui Ding\textsuperscript{\rm 3}\thanks{Corresponding author} \vspace{2mm} \\
         \textsuperscript{\rm 1}~Xiaohongshu Inc. 
         \textsuperscript{\rm 2}~AIM Lab, University of Amsterdam \\
         \textsuperscript{\rm 3}~Institute of Big Data, Fudan University
         \\
        {\tt\small \{jaaackhong, tattoo.ysl, henghui.ding\}@gmail.com}
}

\begin{document}
\maketitle

\begin{abstract}
In this work, we propose a progressive scaling training strategy for visual object tracking, systematically analyzing the influence of training data volume, model size, and input resolution on tracking performance. Our empirical study reveals that while scaling each factor leads to significant improvements in tracking accuracy, naïve training suffers from suboptimal optimization and limited iterative refinement. To address this issue, we introduce DT-Training, a progressive scaling framework that integrates small teacher transfer and dual-branch alignment to maximize model potential. The resulting scaled tracker consistently outperforms state-of-the-art methods across multiple benchmarks, demonstrating strong generalization and transferability of the proposed method. Furthermore, we validate the broader applicability of our approach to additional tasks, underscoring its versatility beyond tracking.
\end{abstract}
\section{Introduction}
Visual object tracking~\cite{bertinetto2016fully,li2019siamrpn++,wu2013online,chen2020siamese} is a fundamental task in computer vision, which involves localizing a target object in each video frame based on the initial bounding box given in the first frame. It has various practical applications, such as self-driving~\cite{zhang2016instance,chen2015deepdriving,geiger2012we}, visual surveillance~\cite{xing2010multiple,tian2005robust}, and video compression~\cite{itti2004automatic}. Recent studies have shown that increasing model size or input resolution can improve tracking performance. However, the computational cost often increases disproportionately compared to the actual performance gains. For example, in the case of OSTrack~\cite{ye2022joint}, when scaling from ViT-Base with a resolution of 256 to ViT-Large with a resolution of 384, the computational burden grows substantially, yet the accuracy improvement is modest with only a 2.4-point increase on the LaSOT benchmark, \ie, from 68.4 to 70.8. The challenge of effectively scaling tracking models to fully leverage their potential remains largely unexplored.

\begin{figure*}[t]
\begin{center}
    \includegraphics[width=0.99\linewidth]{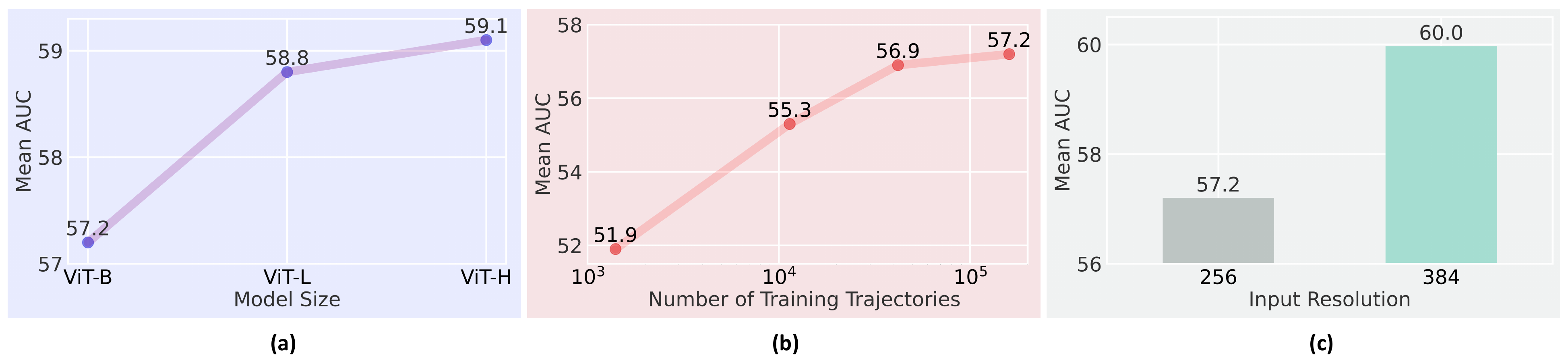}
\end{center}
\vspace{-6mm}
\caption{\textbf{Pioneer Experiments.} We analyze the impact of three key factors in visual object tracking: (a) model size, (b) training data volume, and (c) input resolution.
}
\label{fig:scale_pioneer}
\vspace{-5mm}
\end{figure*}

Thus, we explore scaling strategies to enhance tracking accuracy. We systematically scale model parameters, training data volume, and input resolution, conducting comprehensive experiments to assess their impact. As shown in Figure~\ref{fig:scale_pioneer}, our results reveal a consistent scaling trend, where increasing these factors leads to stable improvements.

Despite the improved accuracy, existing naive training methods encounter several issues based on our observation in Figure~\ref{fig:scale_pioneer}. 1) Directly training large models with extensive data is difficult to optimize and often unstable. 2) Larger models struggle to fully utilize their capacity due to inefficient training dynamics. 3) The open-loop training fails to leverage knowledge gained from previous training. To address this, we introduce a novel progressive scaling approach, DT-Training. In our DT-Training, a smaller model acts as a teacher, guiding the optimization of a larger model for smoother training. Additionally, DT-Training incorporates a dual-branch alignment technique, which applies random masks to input images and aligns outputs from both masked and unmasked images. This increases training difficulty, fully harnessing the model's potential. Through DT-Training, we enable continuous iterative expansion, where the smaller model from the previous iteration transfers knowledge to the larger model. This iterative process transforms scaling into an evolving cycle, consistently enhancing accuracy with each iteration. Our DT-Training achieves a $4.7\%$ improvement on LaSOT when scaling from ViT-Base to ViT-Large at 384 resolution, doubling the gain of naive training ($2.4\%$).

Existing models often evaluate the performance on limited benchmarks that lack the diversity and complexity required to assess robustness in real-world scenarios. Thus, we introduce GTrack Bench, a comprehensive, challenging, and large-scale benchmark featuring 4,369 trajectories, approximately three times the size of existing benchmarks. With our DT-Training, our scaled model shows exceptional capabilities, outperforming current counterparts on GTrack Bench. Our model achieves 64.8 mean AUC, exceeding state-of-the-art methods by at least 1.4 mean AUC. Furthermore, it exhibits strong transferability, maintaining high performance even after compression and proving robust to multimodal data, such as depth maps. By integrating our model into the backbones of CompressTracker~\cite{hong2024general} and OneTracker~\cite{hong2024onetracker}, we achieve consistent performance improvements. Additionally, we also apply our strategy to other downstream vision tasks, such object detection, enhances the accuracy of Deformable DETR~\cite{zhu2020deformable} by 1.5 AP, which demonstrates generalization ability of our method.

Our contributions are summarized as follows: 1) Comprehensive scaling analysis. We investigate the impact of model size, training data volume, and input resolution on visual object tracking. While scaling improves performance, optimization challenges often limit the effectiveness of larger models. 2) Progressive training framework. We propose DT-Training, a novel progressive training approach where a smaller model guides the optimization of a larger one, and outputs from clean and masked images are aligned. This strategy accelerates convergence, stabilizes training, and unlocks the model’s full potential. Additionally, it enables iterative expansion, ensuring that increasing model capacity is effectively utilized across training stages. 3) State-of-the-art performance and generalization. Our scaled model achieves 64.8 mean AUC on GTrack Bench, surpassing existing methods by at least 1.4 mean AUC. Furthermore, experiments on object detection confirm the generalization capability of our approach.

\section{Related Works}
\subsection{Scaling Law in Upstream Tasks}
Scaling laws in neural language processing and vision pretraining tasks have been extensively studied in prior works~\cite{hestness2017deep,sun2017revisiting,brown2020language}. Studies such as~\cite{hoffmann2022training, kaplan2020scaling, tay2021scale, touvron2023llama} explore neural scaling laws in language models, demonstrating a power law relationship between model performance and the scale of model size, data, and training compute. Similar power law dependencies have also been observed in vision tasks~\cite{riquelme2021scaling, zhai2022scaling, dehghani2023scaling,kolesnikov2020big,xie2023data,alabdulmohsin2024getting,fang2023instructseq,yan2025crosslmm,hong2025worldsense,ma2024ee}. Additionally, works like~\cite{radford2021learning,pham2023combined,jia2021scaling,alabdulmohsin2022revisiting,radford2021learning,ramesh2022hierarchical,rombach2022high,cherti2023reproducible,fang2022data,yu2022coca,xiao2025dynaprompt} leverage vast datasets of weakly aligned image-text pairs to strengthen the connection between vision and language tasks.

\subsection{Scaling Law in Downstream Vision Tasks}
Significant attention has been directed towards scaling laws in downstream tasks. Studies like~\cite{liu2024neural, xia2024anygraph} investigate neural scaling laws on graph-based models from both model and data perspectives. SMLPer-X~\cite{cai2024smpler} constructs a large-scale human pose and shape estimation dataset, creating a foundational model. Other studies, like~\cite{minderer2024scaling, tschannen2024image,yan2024panovos,yan2024referred,yan2024sanity} focus on expanding training data size. However, scaling laws in the context of visual object tracking have not been thoroughly explored. In this work, we investigate how scaling affects tracking performance.

\subsection{Visual Object Tracking} 
Visual object tracking aims to locate a target object in each frame based on its initial appearance. Traditional tracking methods~\cite{bertinetto2016fully,li2018high,zhang2020ocean,bhat2019learning,danelljan2019atom,li2019siamrpn++,bolme2010visual,henriques2014high,chen2021transformer,yan2021learning} use a two-stream pipeline to separate feature extraction from relation modeling. Recently, the one-stream pipeline have taken a dominant role~\cite{ye2022joint,cui2022mixformer,cui2024mixformerv2,bai2023artrackv2,wei2023autoregressive,chen2022backbone,chen2023seqtrack,gao2023generalized,zhou2025detrack, zhou2023reading} combining these processes into a unified approach. These one-stream models are primarily built on the vision transformer architecture, which utilizes a series of transformer encoder layers. This design enables more effective relationship modeling between the template and search frame, leading to impressive performance. While previous works enhance model performance by increasing model parameters or input resolution, they have not systematically explored the scaling law in visual object tracking tasks.

\section{Progressive Scaled Visual Object Tracking}
\label{sec:scaling}

\begin{figure*}[t]
\begin{center}
    \includegraphics[width=1\linewidth]{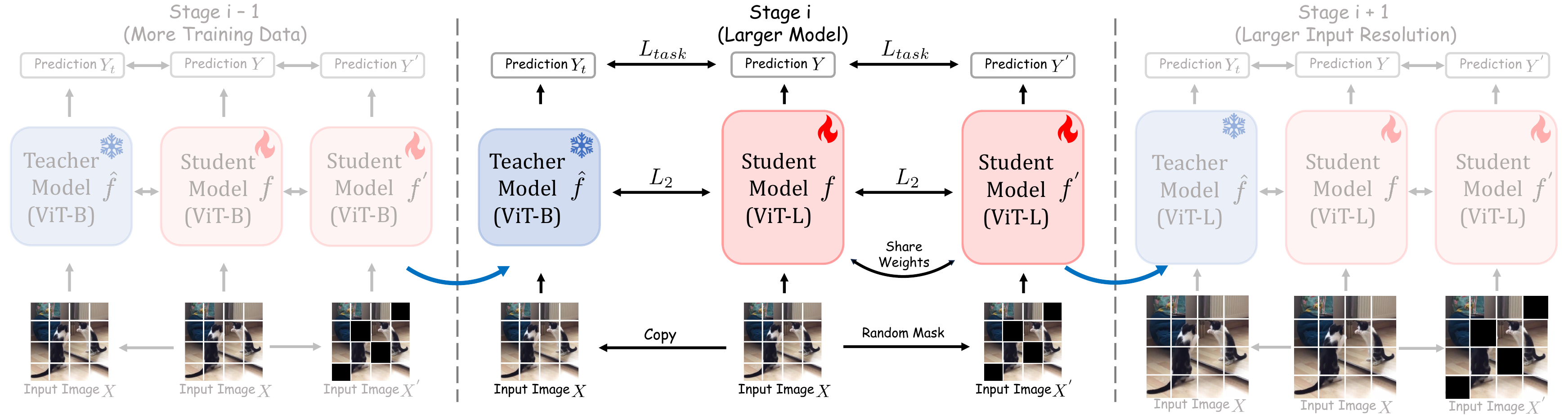}
\end{center}
\vspace{-6mm}
\caption{\textbf{Overview of our progressive scaling approach, DT-Training.} Our DT-Training includes small teacher transfer and dual-branch alignment. We provide an illustrative example of continuous iterative expansion to show a gradual increase in training data, model size, and image resolution. The order of expanding the three key factors is flexible and can be adjusted as needed.
}
\label{fig:training}
\vspace{-3mm}
\end{figure*}

In this section, we first conduct pioneer experiments to investigate the factors that influence visual object tracking performance, focusing on model size, training data volume, and image resolution. We then analyze the limitations of naive training methods, which struggle to fully optimize larger models and fail to leverage the potential of increasing model capacity. To address these issues, we introduce the progressive scaling approach, DT-Training, that guides the training of larger models through a smaller teacher model and incorporates iterative training for progressive scaling. This method enables smoother convergence and better utilization of model potential across successive iterations.

\subsection{Pioneer Experiments}
To investigate the key factors affecting model performance, we adopt OSTrack~\cite{ye2022joint}, which features a ViT~\cite{dosovitskiy2020image} encoder for joint feature extraction and temporal matching, and a lightweight decoder for box regression, for our experiments. This simple architecture allows us to effectively assess the impact of three factors in visual object tracking. As shown in Figure~\ref{fig:scale_pioneer}, by keeping all other variables constant and scaling only one factor at a time, we observe a consistent pattern across all three dimensions: larger models, more extensive training data, and higher input resolutions, each results in improved performance. These observations highlight the critical role of balancing model size, data quantity, and input resolution to optimize visual model performance.

\subsection{Shortcuts of Naive Training}
As shown in preceding pioneer experiments and Figure~\ref{fig:scale_pioneer}, we observe that while expanding certain factors like model size or training data can rapidly enhance model performance up to a specific threshold, beyond the certain point, further expansion results in less noticeable improvements. For example, a model using ViT-H as its backbone only achieves a $0.3 \%$ increase in mean AUC compared to the ViT-L model. Similarly, the performance gains from expanding training data gradually slow down. We attribute these limitations to conventional training approaches. 1) \textbf{Convergence difficulty.} Training a large model directly on extensive datasets can be challenging to optimize due to the increased complexity and computational demands, often leading to issues like slow convergence or getting stuck in local minima. 2) \textbf{Underexplored Capabilities.} Traditional training often fails to fully exploit larger models' capabilities. While these models can capture stronger patterns, conventional training uses fixed training protocols and architectures may hinder their potential, resulting in suboptimal performance. 3) \textbf{Isolate optimization.} Traditional methods follow a linear open-loop process, treating each scaling factor independently, without iterative knowledge sharing. This prevents models from leveraging prior insights, hindering optimization and limiting the full benefits of scaling laws. These limitations underscore the need for a more integrated training approach to maximize model performance.

\subsection{DT-Training}
To address the aforementioned challenges, we introduce a novel progressive scaling approach called DT-Training, as shown in Figure~\ref{fig:training}.  DT-Training integrates dual-branch alignment and small teacher transfer, to fully harness the potential of large models and improve performance. Moreover, DT-Training enables a continuous iterative expansion. In this process, the small model from the previous iteration serves as a teacher to transfer knowledge to the larger model, which then becomes the starting point for the next iteration. This setup facilitates continuous iterative expansion, transforming the scaling process into an evolving cycle that consistently enhances performance.

Directly training large models with excessive parameters often leads to challenges in pattern exploration and optimization difficulty. 
To solve the optimization difficulty problem, we introduce the small teacher transfer approach, where we employ a small pretrained model as a teacher to guide the optimization of the larger model, facilitating smoother learning and faster convergence for the larger model. 
Specifically, in our small teacher transfer, the original images $X$ are simultaneously fed into the training model $f$ and teacher model $\hat{f}$. 
To facilitate the optimization of the student model from different levels, we minimize the distances of both the prediction output and intermediate features.
Given the output $Y$ and intermediate features $F$ obtained by the student model $(Y, F) = f(X)$ and teacher model $(\hat{Y}, \hat{F}) = \hat{f}(X)$, the objective is formulated as:
\begin{equation}
   L_{transfer}(f; \hat{f}) = L_{track}(Y, \hat{Y}) + L_{2}(F,\hat{F}),
   \label{eq: teacher}
\end{equation}
where $L_{2}(F,\hat{F})$ denotes the L2 distance between the features $F$ and $\hat{F}$. $L_{track}(Y, \hat{Y})$ is loss function for tracking. 
Note that we only update the parameters of the student model, and the teacher model is frozen.
With Eq. (\ref{eq: teacher}), our method encourages comprehensive knowledge transfer between teacher and student models, facilitating smoother and more stable optimization for the student model.

To further exploit the ability of the model, we introduce the dual-branch alignment technique, where we apply random masks to input images and align the masked and unmasked image processes. By doing so, we improve the  robustness of the model,
thus unlocking the model's full potential.
Specifically, to introduce additional complexity and promote generalization, we apply random masks to the origin image $X$, generating masked image $X^{'}$. This creates two parallel branches: a clean branch for the original image and a masked branch for the masked image, both of which share the same network weights.
We then obtain the outputs and intermediate features of both the clean image $X$ and masked image $X^{'}$ by the shared student network $f$, formulated as:
\begin{equation}
    (Y, F) = f(X), ~~~ (Y^{'}, F^{'}) = f(X^{'}), 
\end{equation}
where $Y^{'}, F^{'}$ are the predictions and intermediate features from the masked branch, respectively. To optimize the model, we first utilize use groundtruth supervision for the clean branch defined as:
\begin{equation}
   L_{clean}(f) = L_{track}(Y, G), 
\end{equation}
where $L_{clean}$ denotes the task-specific loss for the clean branch and $G$ is the groundtruth label. 
Moreover, similar to Eq. \ref{eq: teacher}, we align the clean and masked student branches by minimizing the distance between both the outputs and intermediate features.
The loss for dual-branch alignment $L_{align}$ is then given by:
\begin{equation}
   L_{align}(f) = L_{track}(Y, Y^{'}) + L_{2}(F,F^{'}). 
\end{equation}

While we use $L_{track}$ to compute the differences between the branches' outputs, more complex methods could also be applied. This loss function is designed to ensure both final predictions and intermediate features from the two branches are aligned, enhancing model’s ability to generalize and leverage its full potential. 

Finally, we combine the dual-branch alignment and small teacher transfer to jointly optimise the model. The overall loss function is formulated as: 
\begin{align}
    L_{total}(f; \hat{f}) = & L_{clean}(f) + \lambda_{transfer}L_{transfer}(f; \hat{f}) \notag \\ & + \lambda_{align}L_{align}(f),
\end{align}    
where $\lambda_{align}$ and $\lambda_{transfer}$ serve as the regularization parameters to balance these components. 
Overall, the knowledge transfer from the teacher to the student model allows the student to leverage the teacher’s pretrained understanding of the task, enabling faster convergence and more efficient learning. Additionally, the masked branch operates with incomplete visual information due to occlusions caused by the random masks. 
This missing local information makes the task more demanding for the masked branch compared to the clean branch. Aligning the two branches enhances the robustness of the student model to incomplete and noisy data, resulting in stronger representational capabilities.
Through the combination of dual-branch alignment and teacher model transfer, we address the optimization difficulty of naive training approaches and further exploit model’s capability.

\begin{table*}[t]
\begin{center}
\caption{\textbf{GTrack Bench statics.} GTrack Bench consists of 12 challenging benchmarks and roughly 4 times the trajectory number provided by current popular benchmarks.}
\label{tab:gtrack_data}
\vspace{-3mm}
\begin{adjustbox}{width=1.\linewidth}
{
\tiny
\begin{tabular}{lccccccccccccccc}
\toprule
& \rot{LaSOT} & \rot{LaSOT$_{ext}$} & \rot{TrackingNet} & \rot{TNL2K} & \rot{UAV123} & \rot{Avist} & \rot{LaGOT} & \rot{LaTOT} & \rot{HOOT} & \rot{VideoCube} & \rot{MOSE} & \rot{OVIS} & \rot{Sum} \\

\midrule
Trajectories & 280 & 150 & 511 & 600 & 123 & 120 & 850 & 165 & 130 & 50 & 531 & 859 & 4369 \\
Videos       & 280 & 150 & 511 & 600 & 123 & 120 & 280 & 165 & 130 & 50 & 200 & 200 & 3379 \\
Mean Frames  & 2512 & 2395 & 441 & 697 & 1247 & 666 & 2512 & 684 & 730 & 14267 & 70 & 78 & - \\
\bottomrule
\end{tabular}

}
\end{adjustbox}
\end{center}
\vspace{-5mm}
\end{table*}

\begin{table*}[t]
\begin{center}
\caption{\textbf{Effectiveness of DT-Training.} We compare the performance between our DT-Training and the conventional training approach under the same conditions. For `Baseline-B-256-N', 'Baseline' indicates model name, `B' refers to ViT-B, `256' specifies the input resolution, and `N' represents training data. N refers to normally used four tracking datasets, and M represents more training data.}
\label{tab:main_all}
\vspace{-3mm}
\begin{adjustbox}{width=1.\linewidth}
{
\renewcommand\arraystretch{1.06}
\begin{tabular}{l|ccc|ccc|ccc|c}
\toprule
\multirow{2}{*}{Model}  & \multicolumn{3}{c|}{LaSOT} & \multicolumn{3}{c|}{LaSOT$_{ext}$} &   \multicolumn{3}{c|}{TNL2K} & Mean  \\
\cline{2-10}
                       & AUC   & P$_{Norm}$ & P    & AUC        & P$_{Norm}$ & P    & AUC   & P$_{Norm}$ & P & AUC    \\
\hline

Baseline-B-256-N         & 68.4  & 77.8  & 74.2 & 47.0       & 57.0 & 52.9 & 56.4  & 71.7  & 58.4 & 57.3 \\
\midrule
\multicolumn{11}{c}{\textit{Training Data Scale Up}} \\
\midrule
Baseline-B-256-M & 68.6  & 78.3  & 74.2 & 47.3       & 55.9  & 51.8 & 60.5  & 76.9  & 65.0 & 58.8 \\
\cellcolor{gray!15}{\textbf{Ours-B-256-M}}     & \cellcolor{gray!15}{\textbf{69.5}}  & \cellcolor{gray!15}{\textbf{79.2}}  &  \cellcolor{gray!15}{\textbf{75.3}}  & \cellcolor{gray!15}{\textbf{47.9}}       & \cellcolor{gray!15}{\textbf{57.5}} & \cellcolor{gray!15}{\textbf{53.5}} & \cellcolor{gray!15}{\textbf{61.2}}  & \cellcolor{gray!15}{\textbf{77.2}}  &  \cellcolor{gray!15}{\textbf{65.0}}   & \cellcolor{gray!15}{\textbf{59.5}} \\
\midrule
\multicolumn{11}{c}{\textit{Model Size Scale Up}} \\
\midrule
Baseline-L-256-N    & 70.0  & 79.2 & 76.3 & 46.6       & 56.9   & 53.0 & 59.6  & 71.9 & 58.9  & 58.7 \\
\cellcolor{gray!15}{\textbf{Ours-L-256-N}}        & \cellcolor{gray!15}{\textbf{71.0}}  & \cellcolor{gray!15}{\textbf{80.9}}   & \cellcolor{gray!15}{\textbf{77.2}} & \cellcolor{gray!15}{\textbf{46.0}}       &  \cellcolor{gray!15}{\textbf{55.9}}  &  \cellcolor{gray!15}{\textbf{52.2}}  & \cellcolor{gray!15}{\textbf{60.1}}  & \cellcolor{gray!15}{\textbf{72.6}}  &  \cellcolor{gray!15}{\textbf{59.5}} & \cellcolor{gray!15}{\textbf{59.2}} \\
\midrule
\multicolumn{11}{c}{\textit{Input Resolution Scale Up}} \\
\midrule
Baseline-B-384-N     & 70.0  & 79.4   &  76.1  & 51.4     & 62.2  & 58.1 & 58.5  &  70.7  & 57.0   & 60.0 \\
\cellcolor{gray!15}{\textbf{Ours-B-384-N}}         & \cellcolor{gray!15}{\textbf{70.6}}  &  \cellcolor{gray!15}{\textbf{80.3}} & \cellcolor{gray!15}{\textbf{76.8}} & \cellcolor{gray!15}{\textbf{51.9}}       &  \cellcolor{gray!15}{\textbf{62.6}} & \cellcolor{gray!15}{\textbf{58.6}} & \cellcolor{gray!15}{\textbf{59.4}}  & \cellcolor{gray!15}{\textbf{72.0}} & \cellcolor{gray!15}{\textbf{58.1}}  & \cellcolor{gray!15}{\textbf{60.6}} \\
\bottomrule
\end{tabular}
}
\end{adjustbox}
\end{center}
\vspace{-6mm}
\end{table*}

\subsection{Progressive Scaling up}
Based on the DT-Training, we can implement the progressive scaling up, which is shown in Figure~\ref{fig:training}. The key idea behind progressive scaling is to progressively increase model size, training data, and input resolution in a controlled manner over multiple iterations. Instead of directly scaling up a large model at the start, we begin with a smaller model and gradually expand its capacity as training progresses. At each iteration, we utilize the model from the previous step as the foundation for the next stage of training. The smaller, previously trained model serves as a guide for optimizing the larger model, allowing us to achieve smoother convergence and avoid the optimization challenges that often arise when training very large models from scratch. Each new iteration introduces an increase in either model parameters, or training data, or input resolution, gradually expanding the model's capacity.

Our DT-Training enables the feasibility of a progressive scaling strategy, offering key advantages over traditional methods. First, the iterative teacher-student relationship allows each new student model to inherit the accumulated knowledge of previous iterations, leading to faster convergence and better generalization. Second, while conventional training often faces diminishing returns as models are scaled, our strategy transforms scaling into an iterative refinement process, ensuring consistent improvement. Additionally, the progressive scaling strategy offers excellent scalability, making it suitable for progressively enlarging models and more complex datasets as the training advances.

\subsection{Training and Inference}
Following previous works~\cite{ye2022joint}, we adopt the the weighted focal loss $L_{cls}$, predicted bounding box $L_{1}$, and the generalized IoU loss $L_{iou}$ for the final loss function, which can be formulated as:
\begin{equation}
    L_{track} = L_{cls} + \lambda_{iou} L_{iou} + \lambda_{L_{1}} L_{1}, 
\end{equation}
where $\lambda_{iou} = 2$ and $\lambda_{L_{1} = 5}$ are the regularization parameters. For inference, we adopt Hanning window penalty to utilize
positional prior in tracking.

\subsection{Discussion}
\textbf{Small Teacher Transfer.} 
we use a smaller model to guide the training of a larger model, a strategy that contrasts with the traditional teacher-student framework commonly used in knowledge distillation. The motivation is to overcome the optimization difficulties that arise when training large models with large datasets. Instead of following the conventional distillation process, where a large teacher model transfers knowledge to a smaller student model, our approach reverses this relationship. Furthermore, our DT-Training enables iterative optimization through small teacher transfer, a dynamic process that traditional knowledge transfer methods cannot achieve. 

\textbf{Scaling Order.} 
The progressive scaling process is flexible, with no strict rules on the order or manner of scaling. At any training stage, we can scale model size, training data volume, or input resolution, individually or jointly, without requiring a predetermined sequence.

\textbf{Inference Cost.}
Another key advantage of our approach is that it does not impact the model's inference speed at test time, as the scaled model preserves the original computational overhead while delivering improved performance.

\section{Experiments}

\begin{table*}[t]
\begin{center}
\caption{\textbf{Effectiveness of progressive scaling up strategy.} Performance comparison with naive training on GTrack Bench.}
\label{tab:main}
\vspace{-3mm}
\begin{adjustbox}{width=1.\linewidth}
{
\begin{tabular}{lccccccccccccc}
\toprule
Model              & LaSOT & LaSOT$_{ext}$ & TrackingNet & TNL2K & UAV123 & Avist & LaGOT & LaTOT & HOOT & VideoCube & MOSE & OVIS &  Mean  \\
\midrule
Baseline-B-256-N     & 68.4  & 47.0       & 83.5        & 56.4  & 67.8   & 57.0   & 61.9  & 28.9  & 56.4 & 45.5 & 51.4 & 55.3 & 59.4 \\
\cellcolor{gray!10}{\textbf{Ours-B-256-M}}   & \cellcolor{gray!10}{\textbf{69.5}}  & \cellcolor{gray!10}{\textbf{47.9}}       & \cellcolor{gray!10}{\textbf{83.6}}        & \cellcolor{gray!10}{\textbf{61.2}}  & \cellcolor{gray!10}{\textbf{69.2}}   & \cellcolor{gray!10}{\textbf{57.6}}  & \cellcolor{gray!10}{\textbf{63.1}}  & \cellcolor{gray!10}{\textbf{30.6}}  & \cellcolor{gray!10}{\textbf{56.5}} & \cellcolor{gray!10}{\textbf{47.4}} & \cellcolor{gray!10}{\textbf{55.5}} & \cellcolor{gray!10}{\textbf{60.1}} & \cellcolor{gray!10}{\textbf{62.0}} \\
\midrule
Baseline-L-256-N     & 70.0  & 46.6       & \textbf{84.4}        & 59.6  & 67.9   & 58.3  & 62.4  & 30.2  & 61.1 & 47.4 & 52.4 & 57.5 & 60.9 \\
\cellcolor{gray!10}{\textbf{Ours-L-256-M}}         & \cellcolor{gray!10}{\textbf{71.6}}  & \cellcolor{gray!10}{\textbf{48.2}}       & \cellcolor{gray!10}{84.2}        & \cellcolor{gray!10}{\textbf{65.0}}  & \cellcolor{gray!10}{\textbf{69.1}}   & \cellcolor{gray!10}{\textbf{60.1}}  & \cellcolor{gray!10}{\textbf{65.2}}  & \cellcolor{gray!10}{\textbf{30.5}}  & \cellcolor{gray!10}{\textbf{62.0}} & \cellcolor{gray!10}{\textbf{48.5}} & \cellcolor{gray!10}{\textbf{55.6}} & \cellcolor{gray!10}{\textbf{61.2}} & \cellcolor{gray!10}{\textbf{63.6}} \\
\midrule
Baseline-L-384-N & 70.8  & 47.0       & \textbf{85.0}        & 60.5  & \textbf{70.3}   & 59.6  & 63.4  & 31.0  & 61.8 & 48.6  & \textbf{57.5} & \textbf{63.3} & 63.4 \\
\cellcolor{gray!10}{\textbf{Ours-L-384-M}}     & \cellcolor{gray!10}{\textbf{73.1}}  & \cellcolor{gray!10}{\textbf{53.0}}       & \cellcolor{gray!10}{84.7}        & \cellcolor{gray!10}{\textbf{66.3}}  & \cellcolor{gray!10}{69.7}   & \cellcolor{gray!10}{\textbf{60.5}}  & \cellcolor{gray!10}{\textbf{67.3}}  & \cellcolor{gray!10}{\textbf{32.0}}  & \cellcolor{gray!10}{\textbf{62.0}} & \cellcolor{gray!10}{\textbf{53.1}} & \cellcolor{gray!10}{55.7} & \cellcolor{gray!10}{61.5} & \cellcolor{gray!10}{\textbf{64.8}} \\
\bottomrule
\end{tabular}
}
\end{adjustbox}
\vspace{-5mm}
\end{center}

\end{table*}

\begin{table*}[t]
\begin{center}
\caption{\textbf{Comparison with state-of-the-art models on GTrack Bench.} Our models significantly outperform state-of-the-art counterparts, highlighting the effectiveness of our progressive scaling up strategy.}
\label{tab:main_compare}\vspace{-3mm}
\begin{adjustbox}{width=1.\linewidth}
{
\begin{tabular}{lccccccccccccc}
\toprule 

Model              & LaSOT & LaSOT$_{ext}$ & TrackingNet & TNL2K & UAV123 & Avist & LaGOT & LaTOT & HOOT & VideoCube & MOSE & OVIS &  Mean  \\

\midrule  
Baseline-B-256-N & 68.4 & 47.0 & 83.5 & 55.9 & 70.7 & 57.0 & 61.9 & 28.9 & 56.4 & 45.5 & 51.4 & 55.3 & 59.4 \\
GRM-Base~\cite{gao2023generalized} & 69.9 & 47.3 & 84.0 & 57.0 & 70.2 & 54.5 & 62.4 & 28.8 & \textbf{56.7} & 45.4 & 52.4 & \textbf{56.7} & 60.2 \\
SeqTrack-Base~\cite{chen2023seqtrack} & 69.9 & 49.5 & 83.3 & 54.9 & 69.2 & 56.8 & \textbf{63.5} & 29.8 & 50.3 & \textbf{48.5} & 49.8 & 54.7 & 59.3 \\
ARTrack-Base~\cite{wei2023autoregressive} & 70.4 & 46.4 & 84.2 & 57.5 & 67.7 & \textbf{59.9} & 62.7 & \textbf{30.8} & 56.2 & 44.4 & 52.4 & 57.7 & 60.6 \\
ARTrackV2-Base~\cite{bai2024artrackv2} & \textbf{71.6} & \textbf{50.8} & \textbf{84.9} & 59.2 & \textbf{69.9} & - & - & - & - & - & - & - & -\\
\cellcolor{gray!15}{\textbf{Ours-B-256-M}} & \cellcolor{gray!15}{69.5}  & \cellcolor{gray!15}{47.9} & \cellcolor{gray!15}{83.6} & \cellcolor{gray!15}{\textbf{61.2}}  & \cellcolor{gray!15}{69.2}   & \cellcolor{gray!15}{57.6}  & \cellcolor{gray!15}{63.1}  & \cellcolor{gray!15}{30.6}  & \cellcolor{gray!15}{56.5} & \cellcolor{gray!15}{47.4} & \cellcolor{gray!15}{\textbf{55.5}} & \cellcolor{gray!15}{\textbf{60.1}} & \cellcolor{gray!15}{\textbf{62.0}} \\
\midrule
Baseline-L-256-N & 69.9 & 47.1 & 84.4 & 59.6 & 67.9 & 58.3 & 62.4 & 30.2 & 61.1 & 47.4 & 52.4 & 57.5 & 60.9 \\
SeqTrack-L~\cite{chen2023seqtrack} & \textbf{72.1} & \textbf{50.5} & \textbf{85.0} & 56.9 & \textbf{69.7} & \textbf{61.1} & \textbf{65.5} & \textbf{31.5} & 51.4 & \textbf{51.2} & 52.8 & 58.2 & 61.7 \\
\cellcolor{gray!15}{\textbf{Ours-L-256-M}}         & \cellcolor{gray!15}{71.6}  & \cellcolor{gray!15}{48.2}       & \cellcolor{gray!15}{84.2}        & \cellcolor{gray!15}{\textbf{65.0}}  & \cellcolor{gray!15}{69.1}  & \cellcolor{gray!15}{60.1}  & \cellcolor{gray!15}{65.2}  & \cellcolor{gray!15}{30.5}  & \cellcolor{gray!15}{\textbf{62.0}} & \cellcolor{gray!15}{48.5} & \cellcolor{gray!15}{\textbf{55.6}} & \cellcolor{gray!15}{\textbf{61.2}} & \cellcolor{gray!15}{\textbf{63.6}} \\
\midrule
Baseline-L-384-N & 70.8 & 47.0 & 85.0 & 60.5 & 70.3 & 59.6 & 63.4 & 31.0 & 61.8 & 48.6 & 57.5 & 63.3 & 63.4 \\
GRM-L320~\citep{gao2023generalized}  & 71.4 & 51.5 & 84.4 & 58.2 & 70.8 & 57.5 & 64.8 & 32.5 & 58.5 & 50.9 & 51.5 & 56.6 & 61.3\\
SeqTrack-L384~\citep{chen2023seqtrack} & 72.5 & 50.7 & 85.5 & 57.8 & 68.5 & 63.1 & 65.6 & 30.8 & 53.2 & 51.8 & 54.3 & 59.8 & 62.4 \\
ARTrack-L384~\citep{wei2023autoregressive} & 73.1 & 52.4 & 85.6 & 61.1 & 69.2 & \textbf{64.5} & 66.2 & \textbf{34.2} & \textbf{63.1} & 43.0 & 55.3 & 61.3  & 63.9 \\
ARTrackV2-L384~\citep{bai2024artrackv2} & 73.6 & 53.4 & \textbf{86.1} & 61.6 & 71.7 & - & - & - & - & - & - & - & -\\
LoRAT-L-378~\cite{lin2024tracking} & \textbf{75.1} & \textbf{56.6} & 85.6 & 62.3 & \textbf{72.5} & - & - & - & - & - & - & - & -\\

\cellcolor{gray!15}{\textbf{Ours-L-384-M}}     & \cellcolor{gray!15}{73.1}  & \cellcolor{gray!15}{53.0}       & \cellcolor{gray!15}{84.7}        & \cellcolor{gray!15}{\textbf{66.3}}  & \cellcolor{gray!15}{69.7}   & \cellcolor{gray!15}{60.5}  & \cellcolor{gray!15}{\textbf{67.3}}  & \cellcolor{gray!15}{32.0}  & \cellcolor{gray!15}{62.0} & \cellcolor{gray!15}{\textbf{53.1}}  & \cellcolor{gray!15}{\textbf{55.7}} & \cellcolor{gray!15}{\textbf{61.5}}  & \cellcolor{gray!15}{\textbf{64.8}} \\

\bottomrule
\end{tabular}
}
\end{adjustbox}
\vspace{-6mm}
\end{center}

\end{table*}

\subsection{Implement Details}
We choose OSTrack\cite{ye2022joint} as our baseline for its simplicity and effectiveness. Training datasets include LaSOT\cite{fan2019lasot}, TrackingNet\cite{muller2018trackingnet}, GOT-10K\cite{huang2019got}, and COCO\cite{lin2014microsoft}, following OSTrack and MixFormerV2\cite{cui2024mixformerv2}. Since these datasets alone are insufficient to fully train a high-capacity tracker, we adapt datasets from multi-object tracking, video object segmentation, and related tasks into a single-object tracking format.
By incorporating a large number of training trajectories, we quadruple the training data, exceeding the size of the original four datasets.

We train the model with AdamW optimizer~\cite{loshchilov2017decoupled}, with a weight decay of $10^{-4}$ and an initial learning rate of $4 \times 10^{-4}$. The total training epochs is 300 with 60K image pairs per epoch and the learning rate is reduced by a factor of 10 after 240 epochs. We employ a batch size of $256$. The search and template images are resized to resolutions of $256\times 256$ and $128\times 128$ resolutions, respectively. We set $\lambda_{align}$ as 0.1. $\lambda_{transfer}$ are set as 0.5 for the first 270 epochs and reduc to 0.0 for the last 30 epochs. The mask ratio is gradually increased from 0.05 to 0.4. We initialize the model with the pretrained parameters from MAE. To maximize the benefit of extensive training data, we employ a balanced sampling strategy to ensure that larger datasets do not overshadow smaller ones.

\subsection{GTrack Bench}
Existing tracking models~\cite{cui2022mixformer,cui2024mixformerv2,ye2022joint,bai2023artrackv2} tend to assess performance on a limited number of benchmarks (about 3-4, covering approximately 1000 trajectories), including TrackingNet~\cite{muller2018trackingnet}, GOT-10K~\cite{huang2019got}, and LaSOT~\cite{fan2019lasot}. However, these datasets offer insufficient diversity, and the videos lack the complexity required to assess model robustness in real-world scenarios. Thus, we introduce a comprehensive and challenging benchmark, called General Track Bench (GTrack Bench), designed to comprehensively evaluate the ability of tracking models in diverse scenes. GTrack Bench consists of 3379 videos from 12 datasets, with a total of 4369 trajectories, roughly 3 times the number provided by current popular benchmarks (around 1000 trajectories). The statistics of these 12 datasets and GTrack Bench are summarized in Table~\ref{tab:gtrack_data}.
These datasets capture complex scenes where target objects frequently experience occlusions, presenting a higher degree of difficulty. We calculate the mean results of each benchmark to serve as the final score. By integrating this diverse range of datasets, GTrack Bench provides a comprehensive and realistic framework for evaluating model performance across varied and challenging environments. We will use GTrack Bench for evaluation in the following experiments. Please see Supplementary Materials for more details about our GTrack Bench.

\begin{table*}[t]
\begin{center}
\caption{\textbf{Compression experiments.} Our model maintains competitive accuracy after compression.}
\label{tab:dstracker_result}
\vspace{-3mm}
\begin{adjustbox}{width=1.\linewidth}
  {
    \begin{tabular}{l|ccc|cc|cc|ccc|cc}
    \toprule
    \multirow{2}{*}{\textbf{Method}} &
    \multicolumn{3}{c|}{\textbf{LaSOT}} &
    \multicolumn{2}{c|}{\textbf{LaSOT$_{ext}$}} &
    \multicolumn{2}{c|}{\textbf{TNL2K}} &
    \multicolumn{3}{c|}{\textbf{TrackingNet}} &
    \multicolumn{2}{c}{\textbf{UAV123}}\\
    & AUC & P$_{Norm}$ & P & AUC & P & AUC & P & AUC & P$_{Norm}$ & P & AUC & P\\
    \hline

    HiT-Base~\citep{kang2023exploring} & 64.6 & 73.3 & 68.1 & 44.1 & - & - & - & 80.0 & 84.4 & 77.3 & 65.6 & - \\
    HiT-Samll~\citep{kang2023exploring} & 60.5 & 68.3 & 61.5 & 40.4 & - & - & - & 77.7 & 81.9 & 73.1 & 63.3 & -\\
    HiT-Tiny~\citep{kang2023exploring} & 54.8 & 60.5 & 52.9 & 35.8 & - & - & - & 74.6 & 78.1 & 68.8 & 53.2 & - \\
    SMAT~\citep{gopal2024separable} & 61.7 & 71.1 & 64.6 & - & - & - & - & 78.6 & 84.2 & 75.6 & 64.3 & 83.9  \\
    MixFormerV2-S~\citep{cui2024mixformerv2} & 60.6 & 69.9 & 60.4 & 43.6 & 46.2 & 48.3 & 43.0 & 75.8 & 81.1 & 70.4 & 65.8 & 86.8 \\

    \hline
    
    CompressTracker-4~\citep{hong2024general} & 66.1  & 75.2 & 70.6 & 45.7  & 50.8 & 53.6  & 52.5 & 82.1  & 87.6 & 80.1 & 67.4  & 88.0  \\

    \cellcolor{gray!15}{\textbf{CompressTracker-4-Ours}} & \cellcolor{gray!15}{\textbf{66.9}}  & \cellcolor{gray!15}{\textbf{76.3}} & \cellcolor{gray!15}{\textbf{71.7}} & \cellcolor{gray!15}{\textbf{46.0}}  & \cellcolor{gray!15}{\textbf{51.4}} & \cellcolor{gray!15}{\textbf{54.8}}  & \cellcolor{gray!15}{\textbf{54.9}} & \cellcolor{gray!15}{\textbf{82.6}}  & \cellcolor{gray!15}{\textbf{87.9}} & \cellcolor{gray!15}{\textbf{80.5}} & \cellcolor{gray!15}{\textbf{67.9}}  & \cellcolor{gray!15}{\textbf{88.3}}  \\

    \bottomrule
    \end{tabular}
    }
\end{adjustbox}
\vspace{-6mm}
\end{center}
\end{table*}

\begin{table}
\begin{center}
\caption{\textbf{Ablation Study on Small Teacher Transfer \& Dual-Branch Alignment.} We investigate the effects of teacher transfer and dual-branch alignment.}
\vspace{-3mm}
\label{tab:ablation_teacher}
\begin{adjustbox}{width=1.\linewidth}
  {
    \begin{tabular}{ccccccc}
\toprule
\#  & Teacher & Mask & LaSOT & LaSOT$_{ext}$ & TNL2K & Mean \\
\midrule
1 &         &      & 68.4  & 47.0         & 56.4  & 57.3 \\
2 & \checkmark       &      & 68.9  & 47.1       & \textbf{56.7}  & 57.6 \\
3 &         & \checkmark    & 69.4  & 47.2       & 56.5  & 57.7 \\
\cellcolor{gray!15}{4} & \cellcolor{gray!15}{\checkmark}       & \cellcolor{gray!15}{\checkmark}    & \cellcolor{gray!15}{\textbf{70.1}}  & \cellcolor{gray!15}{\textbf{47.4}}       & \cellcolor{gray!15}{56.6}  & \cellcolor{gray!15}{\textbf{58.0}}   \\
\bottomrule
\end{tabular}
    }
\end{adjustbox}
\end{center}
\vspace{-6mm}
\end{table}

\subsection{Progressive Scaling Up}

To validate the effectiveness of our progressive scaling strategy, we compare models trained with our approach against those trained using a conventional naive training paradigm.

\textbf{Effectiveness and Generalization of DT-Training.} Firstly, to assess the generalization capability and effectiveness of our DT-Training method, we start with a baseline model trained on a limited set of commonly used datasets (\textit{e.g.} COCO~\cite{lin2014microsoft}, TrackingNet~\cite{muller2018trackingnet}, LaSOT~\cite{fan2019lasot}, and GOT-10k~\cite{huang2019got}), following previous works~\cite{ye2022joint,bai2024artrackv2,cui2022mixformer}. We then independently examine the impact of three critical factors in scaling law: model size, training data, and image resolution, as explored in Section~\ref{sec:scaling}. The results, presented in Table~\ref{tab:main_all}, demonstrate that our DT-Training consistently surpasses traditional training approaches across the three scaling conditions. Specifically, when only the training data was scaled up, we expand the dataset beyond the initial set (\textit{e.g.}, COCO, TrackingNet, LaSOT, GOT-10k) by adding more diverse and larger-scale datasets, which results in a $0.7\%$ increase in the mean AUC score across three datasets compared to naive training. In cases where only the model size is scaled up, we increase the complexity of the model by using a larger architecture, moving from ViT-B to ViT-L. This adjustment yields  a $0.5\%$ increase in the mean AUC score over naive training. Additionally, when the image resolution is increased from 256 to 384, we observe a performance boost of approximately $0.6\%$ in mean accuracy. In summary, our DT-Training demonstrates significant effectiveness, as evidenced by consistent performance improvements across the three scaling conditions compared to traditional training methods.

\textbf{Effectiveness of progressive scaling up strategy.} We conduct experiments to evaluate the effectiveness of our progressive scaling up strategy and results are shown in Table~\ref{tab:main}. We also adopt the baseline model trained on the four limited datasets (\textit{e.g.}, COCO, TrackingNet, LaSOT, GOT-10k) to serve as the start point of our progressive scaling up process. We progressively expand the training process in three stages: first, we enlarge training volume; second, we scale the model size by transitioning from ViT-B to ViT-L; third, we increase the input image resolution from 256 to 384. Besides, we finetune the scaled model on LaSOT for 40 epochs. We compare the result with naive training the baseline model on the four limited datasets by using the GTrack Bench. We record the AUC score of each benchmark and the mean score. Our model share the same inference speed with baseline model. Our model has a performance gain of at least $2\%$ in the average AUC over ten benchmarks over normal training in all different settings. Our training manner not only is proven to be effective when scaling a single element, but also demonstrate strong effectiveness and flexible scalability compared to naive training in progressive scaling experiments.

\textbf{Comparison with existing models.} To further verify the effectiveness of our progressive scaling up strategy, we compare our models with state-of-the-art counterparts on GTrack Bench, as presented in Table~\ref{tab:main_compare}. Our models achieve competitive accuracy, surpassing existing models by at least 1.4 mean AUC. Notably, while existing models such as ARTrack~\cite{bai2024artrackv2}, and SeqTrack~\cite{chen2023seqtrack} rely on complex architectural designs for performance gains, our models obtain superior results with a simpler structure. This underscores the effectiveness of our progressive scaling strategy.

\subsection{Ablation Study}

To verify the effectiveness of our proposed DT-Training, we conduct a comprehensive analysis of its various components, performing detailed exploratory studies. Unless otherwise noted, the following experiments use a ViT-B model trained on four datasets (COCO, TrackingNet, LaSOT, and GOT-10k) as a teacher model to train another ViT-B tracker on the same datasets, for the purpose of eliminating the influence of other factors, such as resolution, training data volume, and model parameter size.

\begin{figure}[t]
\begin{center}
    \includegraphics[width=0.95\linewidth]{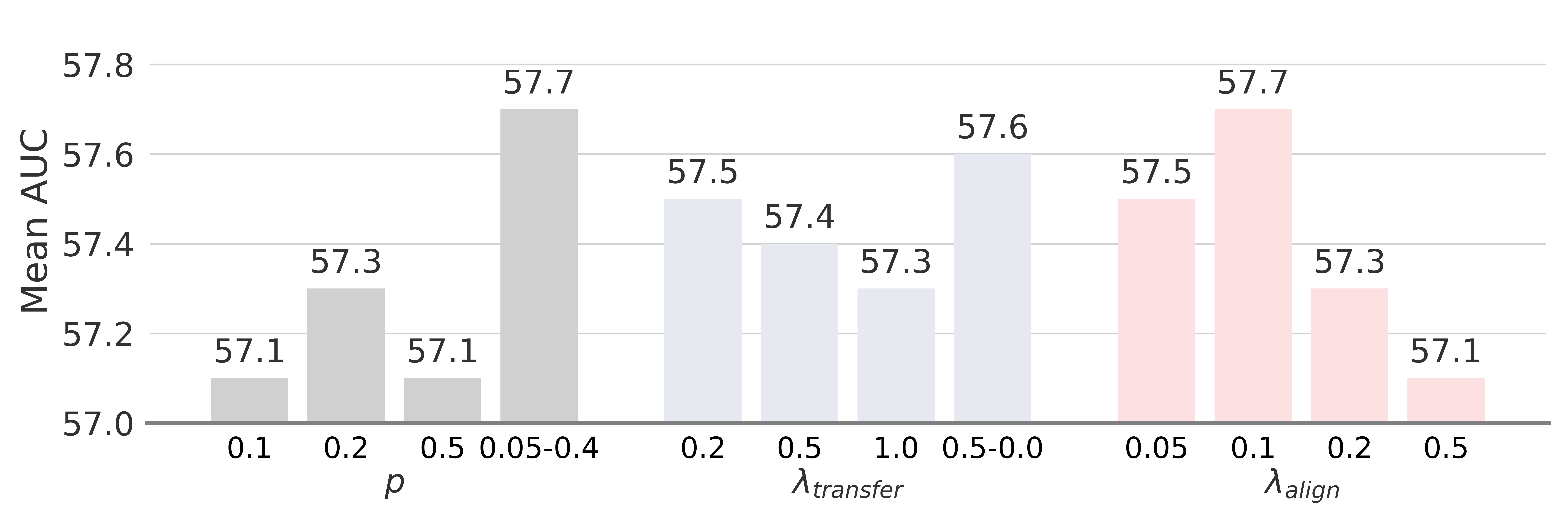}
\end{center}
\vspace{-6mm}
\caption{\textbf{Ablation study on mask ration and regularization parameters.} We conduct experiments to explore the impact of mask ration $p$ and regularization parameters $\lambda_{transfer}$ and $\lambda_{align}$.}
\label{fig:ablation_3}
\end{figure}

\begin{table*}[t]
\begin{center}
\caption{\textbf{Multi-modal robustness experiments.} Our model is robust to multi-modal data.}
\label{tab:onetracker_result}
\vspace{-3mm}
    \resizebox{\textwidth}{!}{
    \tiny
        \begin{tabular}{c  c | ccccc|cc}
    \toprule[.8pt]
    
    &\multicolumn{7}{c}{\textbf{RGB+D Tracking}} & \\
    \hline
    && DeT~\cite{depthtrack} & OSTrack~\cite{ye2022joint} & SPT~\cite{zhu2022rgbd1k} & ProTrack~\cite{protrack} & ViPT~\cite{zhu2023visual} & OneTracker~\cite{hong2024onetracker} & \cellcolor{gray!15}{\textbf{Ours}}\\
    \hline
    \multirow{3}{*}{\tabincell{c}{DepthTrack\\~\cite{yan2021depthtrack}}}&F-score($\uparrow$)&53.2&52.9&53.8&57.8&59.4 & 60.9 & \cellcolor{gray!15}{\textbf{61.6}} \\    
    &R($\uparrow$)&50.6&52.2&54.9&57.3&59.6 & 60.4 & \cellcolor{gray!15}{\textbf{61.2}} \\
    &P($\uparrow$)&56.0&53.6&52.7&58.3&59.2 & 60.7 & \cellcolor{gray!15}{\textbf{61.5}} \\

    \midrule[.8pt]
    
    &\multicolumn{7}{c}{\textbf{RGB+T Tracking}} \\
    \midrule
    &&APFNet~\cite{apfnet} &OSTrack~\cite{ye2022joint} &TransT~\cite{chen2021transformer} &ProTrack~\cite{protrack} & ViPT~\cite{zhu2023visual} & OneTracker~\cite{hong2024onetracker} & \cellcolor{gray!15}{\textbf{Ours}}\\

    \hline
    \multirow{2}{*}{\tabincell{c}{LasHeR\\~\cite{lasher}}}&PR($\uparrow$) &50.0 &51.5 &52.4 &53.8 &65.1 & 67.2 & \cellcolor{gray!15}{\textbf{68.3}} \\
    &SR($\uparrow$) &36.2 &39.4 &41.2 &42.0 &52.5 & 53.8 & \cellcolor{gray!15}{\textbf{55.1}} \\

    \midrule[.8pt]
    
    &\multicolumn{7}{c}{\textbf{RGB+E Tracking}} \\
    \hline
    & & LTMU~\cite{dai2020high} & SiamRCNN~\cite{voigtlaender2020siam} & MDNet~\cite{nam2016learning} & OSTrack~\cite{ye2022joint} & ViPT~\cite{zhu2023visual} & OneTracker~\cite{hong2024onetracker} & \cellcolor{gray!15}{\textbf{Ours}} \\    
    \hline
    \multirow{2}{*}{\tabincell{c}{VisEvent\\~\cite{visevent}}} & MPR($\uparrow$) &65.5 &65.9 &66.1 &69.5 &75.8 & 76.7 & \cellcolor{gray!15}{\textbf{77.4}} \\
    &MSR($\uparrow$) &45.9 &49.9 & - &53.4 &59.2 & 60.8 & \cellcolor{gray!15}{\textbf{61.7}} \\
    \bottomrule[.8pt]

\end{tabular} 
    }
\end{center}
\vspace{-6mm}
\end{table*}
\textbf{Small Teacher Transfer \& Dual-Branch Alignment.}

We conduct experiments to investigate the effects of teacher transfer and dual-branch alignment, with the results presented in Table~\ref{tab:ablation_teacher}. It can be observed that both the small teacher transfer (\# 2) and mask alignment (\# 3) can enhance accuracy compared to naive training (\# 1). Moreover, combining small teacher transfer with mask alignment (\# 4) can further improve model performance. Importantly, by using the same training data, model size, and input image resolution as the baseline training (\# 1), our approach significantly boosts performance, highlighting its effectiveness.

\textbf{Mask Ratio.}
To explore the influence of mask ratio $p$ on mask alignment, we test model performance across different mask ratio and record results on the left side of Figure~\ref{fig:ablation_3}. The results reveal that a low mask ratio (0.1 and 0.2) fails to fully exploit the model's capabilities, while an excessively high mask ratio (0.5) increases training difficulty, negatively impacting performance. Thus, selecting an appropriate mask ratio is crucial to maximizing performance. We begin with a lower mask ratio to allow for faster learning and, as training stabilizes, gradually increase the mask ratio to enhance difficulty, thereby fully harnessing the model's potential (0.05-0.4). This adaptive strategy ensures the model achieves optimal performance by balancing learning ease and training difficulty.

\textbf{Regularization Parameters.}
The regularization parameters also have influence on model performance. As shown in the middle of Figure~\ref{fig:ablation_3}, small teacher transfer enhances model performance, but different $\lambda_{transfer}$ exert a relatively minor influence. In the fourth bar, teacher transfer is employed during the initial 270 epochs to boost training efficiency and performance. In the final 30 epochs, teacher transfer is disabled, allowing the model to independently refine its capabilities, thereby further enhancing performance. This method effectively capitalizes on the strengths of teacher transfer while enabling autonomous learning, resulting in superior model performance. In the right side of Figure~\ref{fig:ablation_3}, we examine the impact of $\lambda_{align}$. We find that both overly high and low $\lambda_{align}$ can negatively impact effectiveness, highlighting the importance of selecting an appropriate $\lambda_{align}$ for optimal results.

\section{Transfer Ability Probing}
In the previous section, we validate the effectiveness of our proposed progressive scaling up strategy, but the transfer ability of our model has not been verified. While our model demonstrates excellent performance across numerous datasets, the transfer ability remains unexplored. Therefore, in this section, we conduct additional experiments to thoroughly evaluate the model's transfer capabilities.

\textbf{Model Compression.} Firstly, we aim to verify whether our model can maintain its excellent performance after compression. We follow CompressTracker~\cite{hong2024general} framework and compress our scaled ViT-B model into a smaller version with just four transformer layers. Except for using a different initial teacher model, all other training parameters, such as data and epochs, remain consistent. As shown in Table~\ref{tab:dstracker_result}, our model achieves superior performance, recording a $66.9 \%$ AUC on LaSOT benchmarks, which is a $0.8\%$ AUC improvement over the original CompressTracker., thanks to our stronger model. Additionally, our model outperforms other lightweight tracking models, confirming its ability to maintain excellent performance after compression.

\textbf{Robustness to multi-modal data.} Furthermore, we investigate the the generalization ability of our model on multimodal data such as thermal maps. By adopting the OneTracker~\cite{hong2024onetracker} architecture,  we explore the adaptability of our models to different modalities, including depth, thermal, and event maps. As shown in Table~\ref{tab:onetracker_result}, our model shows strong generalization to multimodal data. By replacing the backbone of OneTracker~\cite{hong2024onetracker} with our model, OneTracker obtains consistent performance improvement across various multimodal benchmarks. These findings, with our previous experiments, underscore robust transferability of our model.

\begin{table}
\begin{center}
\caption{\textbf{Generalization Experiments.} Our DT-Training can also be applied to other tasks, such as object detection.}
\label{tab:detection_result}
\vspace{-3mm}
    \resizebox{0.49\textwidth}{!}{
        \begin{tabular}{lcccc}
\toprule
Model & AP & AP$_{S}$ & AP$_{M}$ & AP$_{L}$ \\
\midrule
Deformable DETR-R50 & 44.5 & 27.1 & 47.6 & 59.6 \\
\cellcolor{gray!15}{\textbf{Deformable DETR-R50-Ours}} & \cellcolor{gray!15}{\textbf{46.0}} & \cellcolor{gray!15}{\textbf{27.4}} & \cellcolor{gray!15}{\textbf{49.3}} & \cellcolor{gray!15}{\textbf{61.1}} \\

\bottomrule
\end{tabular}
    }
\end{center}
\vspace{-7mm}
\end{table}

\subsection{Generalization Experiments}
Our DT-Training can be applied to other vision tasks. To verify the generalization capability of our method, we conduct experiments on object detection. We apply our method to Deformable DETR~\cite{zhu2020deformable} and train it on COCO~\cite{lin2014microsoft} dataset for 50 epochs, maintaining the original settings. As show in Table~\ref{tab:detection_result}, our method yields a $1.5$ AP performance improvement over origin Deformable DETR under identical settings. Experiments on both tracking and object detection demonstrate that our model effectively operates on both CNN networks and Transformer architectures, demonstrating generalization ability of our method.

\section{Conclusions}
In this work, we explore progressive scaling strategies for visual object tracking, focusing on model size, training data volume, and input resolution. Our analysis reveals that increasing these factors consistently enhances performance. However, training larger models introduces optimization challenges, which we address with DT-Training, a progressive training framework that integrates small teacher transfer and dual-branch alignment. Our approach achieves state-of-the-art performance on the GTrack Bench and demonstrates strong generalization to other tasks, such as object detection. These results underscore the effectiveness and versatility of our method in improving model performance across diverse applications.

{
    \small
    \bibliographystyle{ieeenat_fullname}
    \bibliography{main}

\begin{thebibliography}{104}
\providecommand{\natexlab}[1]{#1}
\providecommand{\url}[1]{\texttt{#1}}
\expandafter\ifx\csname urlstyle\endcsname\relax
  \providecommand{\doi}[1]{doi: #1}\else
  \providecommand{\doi}{doi: \begingroup \urlstyle{rm}\Url}\fi

\bibitem[Achal et~al.(2020)Achal, Tarasha, Pavel, Cordelia, and
  Deva]{achal2020tao}
Dave Achal, Khurana Tarasha, Tokmakov Pavel, Schmid Cordelia, and Ramanan Deva.
\newblock Tao: A large-scale benchmark for tracking any object.
\newblock \emph{European Conference on Computer Vision}, pages 436--454, 2020.

\bibitem[Alabdulmohsin et~al.(2022)Alabdulmohsin, Neyshabur, and
  Zhai]{alabdulmohsin2022revisiting}
Ibrahim~M Alabdulmohsin, Behnam Neyshabur, and Xiaohua Zhai.
\newblock Revisiting neural scaling laws in language and vision.
\newblock \emph{Advances in Neural Information Processing Systems},
  35:\penalty0 22300--22312, 2022.

\bibitem[Alabdulmohsin et~al.(2024)Alabdulmohsin, Zhai, Kolesnikov, and
  Beyer]{alabdulmohsin2024getting}
Ibrahim~M Alabdulmohsin, Xiaohua Zhai, Alexander Kolesnikov, and Lucas Beyer.
\newblock Getting vit in shape: Scaling laws for compute-optimal model design.
\newblock \emph{Advances in Neural Information Processing Systems}, 36, 2024.

\bibitem[Bai et~al.(2023)Bai, Zhao, Gong, and Wei]{bai2023artrackv2}
Yifan Bai, Zeyang Zhao, Yihong Gong, and Xing Wei.
\newblock Artrackv2: Prompting autoregressive tracker where to look and how to
  describe.
\newblock \emph{arXiv preprint arXiv:2312.17133}, 2023.

\bibitem[Bai et~al.(2024)Bai, Zhao, Gong, and Wei]{bai2024artrackv2}
Yifan Bai, Zeyang Zhao, Yihong Gong, and Xing Wei.
\newblock Artrackv2: Prompting autoregressive tracker where to look and how to
  describe.
\newblock In \emph{Proceedings of the IEEE/CVF Conference on Computer Vision
  and Pattern Recognition}, pages 19048--19057, 2024.

\bibitem[Bertinetto et~al.(2016)Bertinetto, Valmadre, Henriques, Vedaldi, and
  Torr]{bertinetto2016fully}
Luca Bertinetto, Jack Valmadre, Joao~F Henriques, Andrea Vedaldi, and Philip~HS
  Torr.
\newblock Fully-convolutional siamese networks for object tracking.
\newblock In \emph{Computer Vision--ECCV 2016 Workshops: Amsterdam, The
  Netherlands, October 8-10 and 15-16, 2016, Proceedings, Part II 14}, pages
  850--865. Springer, 2016.

\bibitem[Bhat et~al.(2019)Bhat, Danelljan, Gool, and Timofte]{bhat2019learning}
Goutam Bhat, Martin Danelljan, Luc~Van Gool, and Radu Timofte.
\newblock Learning discriminative model prediction for tracking.
\newblock In \emph{Proceedings of the IEEE/CVF international conference on
  computer vision}, pages 6182--6191, 2019.

\bibitem[Bolme et~al.(2010)Bolme, Beveridge, Draper, and Lui]{bolme2010visual}
David~S Bolme, J~Ross Beveridge, Bruce~A Draper, and Yui~Man Lui.
\newblock Visual object tracking using adaptive correlation filters.
\newblock In \emph{2010 IEEE computer society conference on computer vision and
  pattern recognition}, pages 2544--2550. IEEE, 2010.

\bibitem[Brown(2020)]{brown2020language}
Tom~B Brown.
\newblock Language models are few-shot learners.
\newblock \emph{arXiv preprint arXiv:2005.14165}, 2020.

\bibitem[Cai et~al.(2024)Cai, Yin, Zeng, Wei, Sun, Yanjun, Pang, Mei, Zhang,
  Zhang, et~al.]{cai2024smpler}
Zhongang Cai, Wanqi Yin, Ailing Zeng, Chen Wei, Qingping Sun, Wang Yanjun,
  Hui~En Pang, Haiyi Mei, Mingyuan Zhang, Lei Zhang, et~al.
\newblock Smpler-x: Scaling up expressive human pose and shape estimation.
\newblock \emph{Advances in Neural Information Processing Systems}, 36, 2024.

\bibitem[Chen et~al.(2022)Chen, Li, Bai, Qiao, Shen, Li, Gan, Wu, and
  Ouyang]{chen2022backbone}
Boyu Chen, Peixia Li, Lei Bai, Lei Qiao, Qiuhong Shen, Bo Li, Weihao Gan, Wei
  Wu, and Wanli Ouyang.
\newblock Backbone is all your need: A simplified architecture for visual
  object tracking.
\newblock In \emph{European Conference on Computer Vision}, pages 375--392.
  Springer, 2022.

\bibitem[Chen et~al.(2015)Chen, Seff, Kornhauser, and
  Xiao]{chen2015deepdriving}
Chenyi Chen, Ari Seff, Alain Kornhauser, and Jianxiong Xiao.
\newblock Deepdriving: Learning affordance for direct perception in autonomous
  driving.
\newblock In \emph{Proceedings of the IEEE international conference on computer
  vision}, pages 2722--2730, 2015.

\bibitem[Chen et~al.(2021)Chen, Yan, Zhu, Wang, Yang, and
  Lu]{chen2021transformer}
Xin Chen, Bin Yan, Jiawen Zhu, Dong Wang, Xiaoyun Yang, and Huchuan Lu.
\newblock Transformer tracking.
\newblock In \emph{Proceedings of the IEEE/CVF conference on computer vision
  and pattern recognition}, pages 8126--8135, 2021.

\bibitem[Chen et~al.(2023)Chen, Peng, Wang, Lu, and Hu]{chen2023seqtrack}
Xin Chen, Houwen Peng, Dong Wang, Huchuan Lu, and Han Hu.
\newblock Seqtrack: Sequence to sequence learning for visual object tracking.
\newblock In \emph{Proceedings of the IEEE/CVF Conference on Computer Vision
  and Pattern Recognition}, pages 14572--14581, 2023.

\bibitem[Chen et~al.(2020)Chen, Zhong, Li, Zhang, and Ji]{chen2020siamese}
Zedu Chen, Bineng Zhong, Guorong Li, Shengping Zhang, and Rongrong Ji.
\newblock Siamese box adaptive network for visual tracking.
\newblock In \emph{Proceedings of the IEEE/CVF conference on computer vision
  and pattern recognition}, pages 6668--6677, 2020.

\bibitem[Cherti et~al.(2023)Cherti, Beaumont, Wightman, Wortsman, Ilharco,
  Gordon, Schuhmann, Schmidt, and Jitsev]{cherti2023reproducible}
Mehdi Cherti, Romain Beaumont, Ross Wightman, Mitchell Wortsman, Gabriel
  Ilharco, Cade Gordon, Christoph Schuhmann, Ludwig Schmidt, and Jenia Jitsev.
\newblock Reproducible scaling laws for contrastive language-image learning.
\newblock In \emph{Proceedings of the IEEE/CVF Conference on Computer Vision
  and Pattern Recognition}, pages 2818--2829, 2023.

\bibitem[Cui et~al.(2022)Cui, Jiang, Wang, and Wu]{cui2022mixformer}
Yutao Cui, Cheng Jiang, Limin Wang, and Gangshan Wu.
\newblock Mixformer: End-to-end tracking with iterative mixed attention.
\newblock In \emph{Proceedings of the IEEE/CVF conference on computer vision
  and pattern recognition}, pages 13608--13618, 2022.

\bibitem[Cui et~al.(2023)Cui, Zeng, Zhao, Yang, Wu, and Wang]{cui2023sportsmot}
Yutao Cui, Chenkai Zeng, Xiaoyu Zhao, Yichun Yang, Gangshan Wu, and Limin Wang.
\newblock Sportsmot: A large multi-object tracking dataset in multiple sports
  scenes.
\newblock In \emph{Proceedings of the IEEE/CVF International Conference on
  Computer Vision}, pages 9921--9931, 2023.

\bibitem[Cui et~al.(2024)Cui, Song, Wu, and Wang]{cui2024mixformerv2}
Yutao Cui, Tianhui Song, Gangshan Wu, and Limin Wang.
\newblock Mixformerv2: Efficient fully transformer tracking.
\newblock \emph{Advances in Neural Information Processing Systems}, 36, 2024.

\bibitem[Dai et~al.(2020)Dai, Zhang, Wang, Li, Lu, and Yang]{dai2020high}
Kenan Dai, Yunhua Zhang, Dong Wang, Jianhua Li, Huchuan Lu, and Xiaoyun Yang.
\newblock High-performance long-term tracking with meta-updater.
\newblock In \emph{Proceedings of the IEEE/CVF conference on computer vision
  and pattern recognition}, pages 6298--6307, 2020.

\bibitem[Danelljan et~al.(2019)Danelljan, Bhat, Khan, and
  Felsberg]{danelljan2019atom}
Martin Danelljan, Goutam Bhat, Fahad~Shahbaz Khan, and Michael Felsberg.
\newblock Atom: Accurate tracking by overlap maximization.
\newblock In \emph{Proceedings of the IEEE/CVF conference on computer vision
  and pattern recognition}, pages 4660--4669, 2019.

\bibitem[Dehghani et~al.(2023)Dehghani, Djolonga, Mustafa, Padlewski, Heek,
  Gilmer, Steiner, Caron, Geirhos, Alabdulmohsin, et~al.]{dehghani2023scaling}
Mostafa Dehghani, Josip Djolonga, Basil Mustafa, Piotr Padlewski, Jonathan
  Heek, Justin Gilmer, Andreas~Peter Steiner, Mathilde Caron, Robert Geirhos,
  Ibrahim Alabdulmohsin, et~al.
\newblock Scaling vision transformers to 22 billion parameters.
\newblock In \emph{International Conference on Machine Learning}, pages
  7480--7512. PMLR, 2023.

\bibitem[Dendorfer(2020)]{dendorfer2020mot20}
P Dendorfer.
\newblock Mot20: A benchmark for multi object tracking in crowded scenes.
\newblock \emph{arXiv preprint arXiv:2003.09003}, 2020.

\bibitem[Dendorfer et~al.(2021)Dendorfer, Osep, Milan, Schindler, Cremers,
  Reid, Roth, and Leal-Taix{\'e}]{dendorfer2021motchallenge}
Patrick Dendorfer, Aljosa Osep, Anton Milan, Konrad Schindler, Daniel Cremers,
  Ian Reid, Stefan Roth, and Laura Leal-Taix{\'e}.
\newblock Motchallenge: A benchmark for single-camera multiple target tracking.
\newblock \emph{International Journal of Computer Vision}, 129:\penalty0
  845--881, 2021.

\bibitem[Ding et~al.(2023)Ding, Liu, He, Jiang, Torr, and Bai]{ding2023mose}
Henghui Ding, Chang Liu, Shuting He, Xudong Jiang, Philip~HS Torr, and Song
  Bai.
\newblock Mose: A new dataset for video object segmentation in complex scenes.
\newblock In \emph{Proceedings of the IEEE/CVF International Conference on
  Computer Vision}, pages 20224--20234, 2023.

\bibitem[Dosovitskiy(2020)]{dosovitskiy2020image}
Alexey Dosovitskiy.
\newblock An image is worth 16x16 words: Transformers for image recognition at
  scale.
\newblock \emph{arXiv preprint arXiv:2010.11929}, 2020.

\bibitem[Du et~al.(2018)Du, Qi, Yu, Yang, Duan, Li, Zhang, Huang, and
  Tian]{du2018unmanned}
Dawei Du, Yuankai Qi, Hongyang Yu, Yifan Yang, Kaiwen Duan, Guorong Li, Weigang
  Zhang, Qingming Huang, and Qi Tian.
\newblock The unmanned aerial vehicle benchmark: Object detection and tracking.
\newblock In \emph{Proceedings of the European conference on computer vision
  (ECCV)}, pages 370--386, 2018.

\bibitem[Fan et~al.(2019)Fan, Lin, Yang, Chu, Deng, Yu, Bai, Xu, Liao, and
  Ling]{fan2019lasot}
Heng Fan, Liting Lin, Fan Yang, Peng Chu, Ge Deng, Sijia Yu, Hexin Bai, Yong
  Xu, Chunyuan Liao, and Haibin Ling.
\newblock Lasot: A high-quality benchmark for large-scale single object
  tracking.
\newblock In \emph{Proceedings of the IEEE/CVF conference on computer vision
  and pattern recognition}, pages 5374--5383, 2019.

\bibitem[Fang et~al.(2022)Fang, Ilharco, Wortsman, Wan, Shankar, Dave, and
  Schmidt]{fang2022data}
Alex Fang, Gabriel Ilharco, Mitchell Wortsman, Yuhao Wan, Vaishaal Shankar,
  Achal Dave, and Ludwig Schmidt.
\newblock Data determines distributional robustness in contrastive language
  image pre-training (clip).
\newblock In \emph{International Conference on Machine Learning}, pages
  6216--6234. PMLR, 2022.

\bibitem[Fang et~al.(2023)Fang, Yan, Huang, Zhou, Tian, Dai, and
  Li]{fang2023instructseq}
Rongyao Fang, Shilin Yan, Zhaoyang Huang, Jingqiu Zhou, Hao Tian, Jifeng Dai,
  and Hongsheng Li.
\newblock Instructseq: Unifying vision tasks with instruction-conditioned
  multi-modal sequence generation.
\newblock \emph{arXiv preprint arXiv:2311.18835}, 2023.

\bibitem[Gao et~al.(2023)Gao, Zhou, and Zhang]{gao2023generalized}
Shenyuan Gao, Chunluan Zhou, and Jun Zhang.
\newblock Generalized relation modeling for transformer tracking.
\newblock In \emph{Proceedings of the IEEE/CVF Conference on Computer Vision
  and Pattern Recognition}, pages 18686--18695, 2023.

\bibitem[Geiger et~al.(2012)Geiger, Lenz, and Urtasun]{geiger2012we}
Andreas Geiger, Philip Lenz, and Raquel Urtasun.
\newblock Are we ready for autonomous driving? the kitti vision benchmark
  suite.
\newblock In \emph{2012 IEEE conference on computer vision and pattern
  recognition}, pages 3354--3361. IEEE, 2012.

\bibitem[Gopal and Amer(2024)]{gopal2024separable}
Goutam~Yelluru Gopal and Maria~A Amer.
\newblock Separable self and mixed attention transformers for efficient object
  tracking.
\newblock In \emph{Proceedings of the IEEE/CVF Winter Conference on
  Applications of Computer Vision}, pages 6708--6717, 2024.

\bibitem[Henriques et~al.(2014)Henriques, Caseiro, Martins, and
  Batista]{henriques2014high}
Jo{\~a}o~F Henriques, Rui Caseiro, Pedro Martins, and Jorge Batista.
\newblock High-speed tracking with kernelized correlation filters.
\newblock \emph{IEEE transactions on pattern analysis and machine
  intelligence}, 37\penalty0 (3):\penalty0 583--596, 2014.

\bibitem[Hestness et~al.(2017)Hestness, Narang, Ardalani, Diamos, Jun,
  Kianinejad, Patwary, Yang, and Zhou]{hestness2017deep}
Joel Hestness, Sharan Narang, Newsha Ardalani, Gregory Diamos, Heewoo Jun,
  Hassan Kianinejad, Md~Mostofa~Ali Patwary, Yang Yang, and Yanqi Zhou.
\newblock Deep learning scaling is predictable, empirically.
\newblock \emph{arXiv preprint arXiv:1712.00409}, 2017.

\bibitem[Hoffmann et~al.(2022)Hoffmann, Borgeaud, Mensch, Buchatskaya, Cai,
  Rutherford, Casas, Hendricks, Welbl, Clark, et~al.]{hoffmann2022training}
Jordan Hoffmann, Sebastian Borgeaud, Arthur Mensch, Elena Buchatskaya, Trevor
  Cai, Eliza Rutherford, Diego de~Las Casas, Lisa~Anne Hendricks, Johannes
  Welbl, Aidan Clark, et~al.
\newblock Training compute-optimal large language models.
\newblock \emph{arXiv preprint arXiv:2203.15556}, 2022.

\bibitem[Hong et~al.(2025)Hong, Yan, Cai, Jiang, Hu, and
  Xie]{hong2025worldsense}
Jack Hong, Shilin Yan, Jiayin Cai, Xiaolong Jiang, Yao Hu, and Weidi Xie.
\newblock Worldsense: Evaluating real-world omnimodal understanding for
  multimodal llms.
\newblock \emph{arXiv preprint arXiv:2502.04326}, 2025.

\bibitem[Hong et~al.(2024{\natexlab{a}})Hong, Li, Zhou, Yan, Guo, Jiang, Chen,
  Gao, Zhang, Lu, et~al.]{hong2024general}
Lingyi Hong, Jinglun Li, Xinyu Zhou, Shilin Yan, Pinxue Guo, Kaixun Jiang,
  Zhaoyu Chen, Shuyong Gao, Wei Zhang, Hong Lu, et~al.
\newblock General compression framework for efficient transformer object
  tracking.
\newblock \emph{arXiv preprint arXiv:2409.17564}, 2024{\natexlab{a}}.

\bibitem[Hong et~al.(2024{\natexlab{b}})Hong, Yan, Zhang, Li, Zhou, Guo, Jiang,
  Chen, Li, Chen, et~al.]{hong2024onetracker}
Lingyi Hong, Shilin Yan, Renrui Zhang, Wanyun Li, Xinyu Zhou, Pinxue Guo,
  Kaixun Jiang, Yiting Chen, Jinglun Li, Zhaoyu Chen, et~al.
\newblock Onetracker: Unifying visual object tracking with foundation models
  and efficient tuning.
\newblock In \emph{Proceedings of the IEEE/CVF Conference on Computer Vision
  and Pattern Recognition}, pages 19079--19091, 2024{\natexlab{b}}.

\bibitem[Hu et~al.(2022)Hu, Zhao, Huang, and Huang]{hu2022global}
Shiyu Hu, Xin Zhao, Lianghua Huang, and Kaiqi Huang.
\newblock Global instance tracking: Locating target more like humans.
\newblock \emph{IEEE Transactions on Pattern Analysis and Machine
  Intelligence}, 45\penalty0 (1):\penalty0 576--592, 2022.

\bibitem[Huang et~al.(2019)Huang, Zhao, and Huang]{huang2019got}
Lianghua Huang, Xin Zhao, and Kaiqi Huang.
\newblock Got-10k: A large high-diversity benchmark for generic object tracking
  in the wild.
\newblock \emph{IEEE transactions on pattern analysis and machine
  intelligence}, 43\penalty0 (5):\penalty0 1562--1577, 2019.

\bibitem[Itti(2004)]{itti2004automatic}
Laurent Itti.
\newblock Automatic foveation for video compression using a neurobiological
  model of visual attention.
\newblock \emph{IEEE transactions on image processing}, 13\penalty0
  (10):\penalty0 1304--1318, 2004.

\bibitem[Jia et~al.(2021)Jia, Yang, Xia, Chen, Parekh, Pham, Le, Sung, Li, and
  Duerig]{jia2021scaling}
Chao Jia, Yinfei Yang, Ye Xia, Yi-Ting Chen, Zarana Parekh, Hieu Pham, Quoc Le,
  Yun-Hsuan Sung, Zhen Li, and Tom Duerig.
\newblock Scaling up visual and vision-language representation learning with
  noisy text supervision.
\newblock In \emph{International conference on machine learning}, pages
  4904--4916. PMLR, 2021.

\bibitem[Kang et~al.(2023)Kang, Chen, Wang, Peng, and Lu]{kang2023exploring}
Ben Kang, Xin Chen, Dong Wang, Houwen Peng, and Huchuan Lu.
\newblock Exploring lightweight hierarchical vision transformers for efficient
  visual tracking.
\newblock In \emph{Proceedings of the IEEE/CVF International Conference on
  Computer Vision}, pages 9612--9621, 2023.

\bibitem[Kaplan et~al.(2020)Kaplan, McCandlish, Henighan, Brown, Chess, Child,
  Gray, Radford, Wu, and Amodei]{kaplan2020scaling}
Jared Kaplan, Sam McCandlish, Tom Henighan, Tom~B Brown, Benjamin Chess, Rewon
  Child, Scott Gray, Alec Radford, Jeffrey Wu, and Dario Amodei.
\newblock Scaling laws for neural language models.
\newblock \emph{arXiv preprint arXiv:2001.08361}, 2020.

\bibitem[Kolesnikov et~al.(2020)Kolesnikov, Beyer, Zhai, Puigcerver, Yung,
  Gelly, and Houlsby]{kolesnikov2020big}
Alexander Kolesnikov, Lucas Beyer, Xiaohua Zhai, Joan Puigcerver, Jessica Yung,
  Sylvain Gelly, and Neil Houlsby.
\newblock Big transfer (bit): General visual representation learning.
\newblock In \emph{Computer Vision--ECCV 2020: 16th European Conference,
  Glasgow, UK, August 23--28, 2020, Proceedings, Part V 16}, pages 491--507.
  Springer, 2020.

\bibitem[Li et~al.(2018)Li, Yan, Wu, Zhu, and Hu]{li2018high}
Bo Li, Junjie Yan, Wei Wu, Zheng Zhu, and Xiaolin Hu.
\newblock High performance visual tracking with siamese region proposal
  network.
\newblock In \emph{Proceedings of the IEEE conference on computer vision and
  pattern recognition}, pages 8971--8980, 2018.

\bibitem[Li et~al.(2019)Li, Wu, Wang, Zhang, Xing, and Yan]{li2019siamrpn++}
Bo Li, Wei Wu, Qiang Wang, Fangyi Zhang, Junliang Xing, and Junjie Yan.
\newblock Siamrpn++: Evolution of siamese visual tracking with very deep
  networks.
\newblock In \emph{Proceedings of the IEEE/CVF conference on computer vision
  and pattern recognition}, pages 4282--4291, 2019.

\bibitem[Li et~al.(2021)Li, Xue, Jia, Qu, Luo, Tang, and Sun]{lasher}
Chenglong Li, Wanlin Xue, Yaqing Jia, Zhichen Qu, Bin Luo, Jin Tang, and Dengdi
  Sun.
\newblock Lasher: A large-scale high-diversity benchmark for {RGBT} tracking.
\newblock \emph{IEEE Transactions on Image Processing}, 31:\penalty0 392--404,
  2021.

\bibitem[Lin et~al.(2024)Lin, Fan, Zhang, Wang, Xu, and Ling]{lin2024tracking}
Liting Lin, Heng Fan, Zhipeng Zhang, Yaowei Wang, Yong Xu, and Haibin Ling.
\newblock Tracking meets lora: Faster training, larger model, stronger
  performance.
\newblock In \emph{European Conference on Computer Vision}, pages 300--318.
  Springer, 2024.

\bibitem[Lin et~al.(2014)Lin, Maire, Belongie, Hays, Perona, Ramanan,
  Doll{\'a}r, and Zitnick]{lin2014microsoft}
Tsung-Yi Lin, Michael Maire, Serge Belongie, James Hays, Pietro Perona, Deva
  Ramanan, Piotr Doll{\'a}r, and C~Lawrence Zitnick.
\newblock Microsoft coco: Common objects in context.
\newblock In \emph{Computer Vision--ECCV 2014: 13th European Conference,
  Zurich, Switzerland, September 6-12, 2014, Proceedings, Part V 13}, pages
  740--755. Springer, 2014.

\bibitem[Liu et~al.(2024)Liu, Mao, Chen, Zhao, Shah, and Tang]{liu2024neural}
Jingzhe Liu, Haitao Mao, Zhikai Chen, Tong Zhao, Neil Shah, and Jiliang Tang.
\newblock Neural scaling laws on graphs.
\newblock \emph{arXiv preprint arXiv:2402.02054}, 2024.

\bibitem[Loshchilov and Hutter(2017)]{loshchilov2017decoupled}
Ilya Loshchilov and Frank Hutter.
\newblock Decoupled weight decay regularization.
\newblock \emph{arXiv preprint arXiv:1711.05101}, 2017.

\bibitem[Ma et~al.(2024)Ma, Zhou, Zhang, Yan, Li, He, Wu, Rao, Zhang, and
  Sun]{ma2024ee}
Feipeng Ma, Yizhou Zhou, Zheyu Zhang, Shilin Yan, Hebei Li, Zilong He, Siying
  Wu, Fengyun Rao, Yueyi Zhang, and Xiaoyan Sun.
\newblock Ee-mllm: A data-efficient and compute-efficient multimodal large
  language model.
\newblock \emph{arXiv preprint arXiv:2408.11795}, 2024.

\bibitem[Mayer et~al.(2024)Mayer, Danelljan, Yang, Ferrari, Van~Gool, and
  Kuznetsova]{mayer2024beyond}
Christoph Mayer, Martin Danelljan, Ming-Hsuan Yang, Vittorio Ferrari, Luc
  Van~Gool, and Alina Kuznetsova.
\newblock Beyond sot: Tracking multiple generic objects at once.
\newblock In \emph{Proceedings of the IEEE/CVF Winter Conference on
  Applications of Computer Vision}, pages 6826--6836, 2024.

\bibitem[Milan et~al.(2016)Milan, Leal-Taix{\'e}, Reid, Roth, and
  Schindler]{milan2016mot16}
Anton Milan, Laura Leal-Taix{\'e}, Ian Reid, Stefan Roth, and Konrad Schindler.
\newblock Mot16: A benchmark for multi-object tracking.
\newblock \emph{arXiv preprint arXiv:1603.00831}, 2016.

\bibitem[Minderer et~al.(2024)Minderer, Gritsenko, and
  Houlsby]{minderer2024scaling}
Matthias Minderer, Alexey Gritsenko, and Neil Houlsby.
\newblock Scaling open-vocabulary object detection.
\newblock \emph{Advances in Neural Information Processing Systems}, 36, 2024.

\bibitem[Mueller et~al.(2016)Mueller, Smith, and Ghanem]{mueller2016benchmark}
Matthias Mueller, Neil Smith, and Bernard Ghanem.
\newblock A benchmark and simulator for uav tracking.
\newblock In \emph{Computer Vision--ECCV 2016: 14th European Conference,
  Amsterdam, The Netherlands, October 11--14, 2016, Proceedings, Part I 14},
  pages 445--461. Springer, 2016.

\bibitem[Muller et~al.(2018)Muller, Bibi, Giancola, Alsubaihi, and
  Ghanem]{muller2018trackingnet}
Matthias Muller, Adel Bibi, Silvio Giancola, Salman Alsubaihi, and Bernard
  Ghanem.
\newblock Trackingnet: A large-scale dataset and benchmark for object tracking
  in the wild.
\newblock In \emph{Proceedings of the European conference on computer vision
  (ECCV)}, pages 300--317, 2018.

\bibitem[Nam and Han(2016)]{nam2016learning}
Hyeonseob Nam and Bohyung Han.
\newblock Learning multi-domain convolutional neural networks for visual
  tracking.
\newblock In \emph{Proceedings of the IEEE conference on computer vision and
  pattern recognition}, pages 4293--4302, 2016.

\bibitem[Noman et~al.(2022)Noman, Ghallabi, Najiha, Mayer, Dudhane, Danelljan,
  Cholakkal, Khan, Van~Gool, and Khan]{noman2022avist}
Mubashir Noman, Wafa~Al Ghallabi, Daniya Najiha, Christoph Mayer, Akshay
  Dudhane, Martin Danelljan, Hisham Cholakkal, Salman Khan, Luc Van~Gool, and
  Fahad~Shahbaz Khan.
\newblock Avist: A benchmark for visual object tracking in adverse visibility.
\newblock \emph{arXiv preprint arXiv:2208.06888}, 2022.

\bibitem[Pham et~al.(2023)Pham, Dai, Ghiasi, Kawaguchi, Liu, Yu, Yu, Chen,
  Luong, Wu, et~al.]{pham2023combined}
Hieu Pham, Zihang Dai, Golnaz Ghiasi, Kenji Kawaguchi, Hanxiao Liu, Adams~Wei
  Yu, Jiahui Yu, Yi-Ting Chen, Minh-Thang Luong, Yonghui Wu, et~al.
\newblock Combined scaling for zero-shot transfer learning.
\newblock \emph{Neurocomputing}, 555:\penalty0 126658, 2023.

\bibitem[Qi et~al.(2022)Qi, Gao, Hu, Wang, Liu, Bai, Belongie, Yuille, Torr,
  and Bai]{qi2022occluded}
Jiyang Qi, Yan Gao, Yao Hu, Xinggang Wang, Xiaoyu Liu, Xiang Bai, Serge
  Belongie, Alan Yuille, Philip~HS Torr, and Song Bai.
\newblock Occluded video instance segmentation: A benchmark.
\newblock \emph{International Journal of Computer Vision}, 130\penalty0
  (8):\penalty0 2022--2039, 2022.

\bibitem[Radford et~al.(2021)Radford, Kim, Hallacy, Ramesh, Goh, Agarwal,
  Sastry, Askell, Mishkin, Clark, et~al.]{radford2021learning}
Alec Radford, Jong~Wook Kim, Chris Hallacy, Aditya Ramesh, Gabriel Goh,
  Sandhini Agarwal, Girish Sastry, Amanda Askell, Pamela Mishkin, Jack Clark,
  et~al.
\newblock Learning transferable visual models from natural language
  supervision.
\newblock In \emph{International conference on machine learning}, pages
  8748--8763. PMLR, 2021.

\bibitem[Ramesh et~al.(2022)Ramesh, Dhariwal, Nichol, Chu, and
  Chen]{ramesh2022hierarchical}
Aditya Ramesh, Prafulla Dhariwal, Alex Nichol, Casey Chu, and Mark Chen.
\newblock Hierarchical text-conditional image generation with clip latents.
\newblock \emph{arXiv preprint arXiv:2204.06125}, 1\penalty0 (2):\penalty0 3,
  2022.

\bibitem[Riquelme et~al.(2021)Riquelme, Puigcerver, Mustafa, Neumann, Jenatton,
  Susano~Pinto, Keysers, and Houlsby]{riquelme2021scaling}
Carlos Riquelme, Joan Puigcerver, Basil Mustafa, Maxim Neumann, Rodolphe
  Jenatton, Andr{\'e} Susano~Pinto, Daniel Keysers, and Neil Houlsby.
\newblock Scaling vision with sparse mixture of experts.
\newblock \emph{Advances in Neural Information Processing Systems},
  34:\penalty0 8583--8595, 2021.

\bibitem[Rombach et~al.(2022)Rombach, Blattmann, Lorenz, Esser, and
  Ommer]{rombach2022high}
Robin Rombach, Andreas Blattmann, Dominik Lorenz, Patrick Esser, and Bj{\"o}rn
  Ommer.
\newblock High-resolution image synthesis with latent diffusion models.
\newblock In \emph{Proceedings of the IEEE/CVF conference on computer vision
  and pattern recognition}, pages 10684--10695, 2022.

\bibitem[Sahin and Itti(2023)]{sahin2023hoot}
Gozde Sahin and Laurent Itti.
\newblock Hoot: Heavy occlusions in object tracking benchmark.
\newblock In \emph{Proceedings of the IEEE/CVF Winter Conference on
  Applications of Computer Vision}, pages 4830--4839, 2023.

\bibitem[Sun et~al.(2017)Sun, Shrivastava, Singh, and Gupta]{sun2017revisiting}
Chen Sun, Abhinav Shrivastava, Saurabh Singh, and Abhinav Gupta.
\newblock Revisiting unreasonable effectiveness of data in deep learning era.
\newblock In \emph{Proceedings of the IEEE international conference on computer
  vision}, pages 843--852, 2017.

\bibitem[Sun et~al.(2022)Sun, Cao, Jiang, Yuan, Bai, Kitani, and
  Luo]{sun2022dancetrack}
Peize Sun, Jinkun Cao, Yi Jiang, Zehuan Yuan, Song Bai, Kris Kitani, and Ping
  Luo.
\newblock Dancetrack: Multi-object tracking in uniform appearance and diverse
  motion.
\newblock In \emph{Proceedings of the IEEE/CVF Conference on Computer Vision
  and Pattern Recognition}, pages 20993--21002, 2022.

\bibitem[Tay et~al.(2021)Tay, Dehghani, Rao, Fedus, Abnar, Chung, Narang,
  Yogatama, Vaswani, and Metzler]{tay2021scale}
Yi Tay, Mostafa Dehghani, Jinfeng Rao, William Fedus, Samira Abnar, Hyung~Won
  Chung, Sharan Narang, Dani Yogatama, Ashish Vaswani, and Donald Metzler.
\newblock Scale efficiently: Insights from pre-training and fine-tuning
  transformers.
\newblock \emph{arXiv preprint arXiv:2109.10686}, 2021.

\bibitem[Tian et~al.(2005)Tian, Lu, and Hampapur]{tian2005robust}
Ying-Li Tian, Max Lu, and Arun Hampapur.
\newblock Robust and efficient foreground analysis for real-time video
  surveillance.
\newblock In \emph{2005 IEEE Computer Society Conference on Computer Vision and
  Pattern Recognition (CVPR'05)}, pages 1182--1187. IEEE, 2005.

\bibitem[Touvron et~al.(2023)Touvron, Lavril, Izacard, Martinet, Lachaux,
  Lacroix, Rozi{\`e}re, Goyal, Hambro, Azhar, et~al.]{touvron2023llama}
Hugo Touvron, Thibaut Lavril, Gautier Izacard, Xavier Martinet, Marie-Anne
  Lachaux, Timoth{\'e}e Lacroix, Baptiste Rozi{\`e}re, Naman Goyal, Eric
  Hambro, Faisal Azhar, et~al.
\newblock Llama: Open and efficient foundation language models.
\newblock \emph{arXiv preprint arXiv:2302.13971}, 2023.

\bibitem[Tschannen et~al.(2024)Tschannen, Kumar, Steiner, Zhai, Houlsby, and
  Beyer]{tschannen2024image}
Michael Tschannen, Manoj Kumar, Andreas Steiner, Xiaohua Zhai, Neil Houlsby,
  and Lucas Beyer.
\newblock Image captioners are scalable vision learners too.
\newblock \emph{Advances in Neural Information Processing Systems}, 36, 2024.

\bibitem[Voigtlaender et~al.(2020)Voigtlaender, Luiten, Torr, and
  Leibe]{voigtlaender2020siam}
Paul Voigtlaender, Jonathon Luiten, Philip~HS Torr, and Bastian Leibe.
\newblock Siam r-cnn: Visual tracking by re-detection.
\newblock In \emph{Proceedings of the IEEE/CVF conference on computer vision
  and pattern recognition}, pages 6578--6588, 2020.

\bibitem[Wang et~al.(2021{\natexlab{a}})Wang, Feiszli, Wang, and
  Tran]{wang2021unidentified}
Weiyao Wang, Matt Feiszli, Heng Wang, and Du Tran.
\newblock Unidentified video objects: A benchmark for dense, open-world
  segmentation.
\newblock In \emph{Proceedings of the IEEE/CVF International Conference on
  Computer Vision}, pages 10776--10785, 2021{\natexlab{a}}.

\bibitem[Wang et~al.(2021{\natexlab{b}})Wang, Li, Zhu, Zhang, Chen, Li, Wang,
  Tian, and Wu]{visevent}
Xiao Wang, Jianing Li, Lin Zhu, Zhipeng Zhang, Zhe Chen, Xin Li, Yaowei Wang,
  Yonghong Tian, and Feng Wu.
\newblock Visevent: Reliable object tracking via collaboration of frame and
  event flows.
\newblock \emph{arXiv preprint arXiv:2108.05015}, 2021{\natexlab{b}}.

\bibitem[Wang et~al.(2021{\natexlab{c}})Wang, Shu, Zhang, Jiang, Wang, Tian,
  and Wu]{wang2021towards}
Xiao Wang, Xiujun Shu, Zhipeng Zhang, Bo Jiang, Yaowei Wang, Yonghong Tian, and
  Feng Wu.
\newblock Towards more flexible and accurate object tracking with natural
  language: Algorithms and benchmark.
\newblock In \emph{Proceedings of the IEEE/CVF conference on computer vision
  and pattern recognition}, pages 13763--13773, 2021{\natexlab{c}}.

\bibitem[Wei et~al.(2023)Wei, Bai, Zheng, Shi, and Gong]{wei2023autoregressive}
Xing Wei, Yifan Bai, Yongchao Zheng, Dahu Shi, and Yihong Gong.
\newblock Autoregressive visual tracking.
\newblock In \emph{Proceedings of the IEEE/CVF Conference on Computer Vision
  and Pattern Recognition}, pages 9697--9706, 2023.

\bibitem[Wu et~al.(2013)Wu, Lim, and Yang]{wu2013online}
Yi Wu, Jongwoo Lim, and Ming-Hsuan Yang.
\newblock Online object tracking: A benchmark.
\newblock In \emph{Proceedings of the IEEE conference on computer vision and
  pattern recognition}, pages 2411--2418, 2013.

\bibitem[Xia and Huang(2024)]{xia2024anygraph}
Lianghao Xia and Chao Huang.
\newblock Anygraph: Graph foundation model in the wild.
\newblock \emph{arXiv preprint arXiv:2408.10700}, 2024.

\bibitem[Xiao et~al.(2022)Xiao, Yang, Li, Liu, and Tang]{apfnet}
Yun Xiao, Mengmeng Yang, Chenglong Li, Lei Liu, and Jin Tang.
\newblock Attribute-based progressive fusion network for {RGBT} tracking.
\newblock In \emph{AAAI}, 2022.

\bibitem[Xiao et~al.(2025)Xiao, Yan, Hong, Cai, Jiang, Hu, Shen, Wang, and
  Snoek]{xiao2025dynaprompt}
Zehao Xiao, Shilin Yan, Jack Hong, Jiayin Cai, Xiaolong Jiang, Yao Hu, Jiayi
  Shen, Qi Wang, and Cees~GM Snoek.
\newblock Dynaprompt: Dynamic test-time prompt tuning.
\newblock \emph{arXiv preprint arXiv:2501.16404}, 2025.

\bibitem[Xie et~al.(2023)Xie, Zhang, Cao, Lin, Wei, Dai, and Hu]{xie2023data}
Zhenda Xie, Zheng Zhang, Yue Cao, Yutong Lin, Yixuan Wei, Qi Dai, and Han Hu.
\newblock On data scaling in masked image modeling.
\newblock In \emph{Proceedings of the IEEE/CVF Conference on Computer Vision
  and Pattern Recognition}, pages 10365--10374, 2023.

\bibitem[Xing et~al.(2010)Xing, Ai, and Lao]{xing2010multiple}
Junliang Xing, Haizhou Ai, and Shihong Lao.
\newblock Multiple human tracking based on multi-view upper-body detection and
  discriminative learning.
\newblock In \emph{2010 20th International Conference on Pattern Recognition},
  pages 1698--1701. IEEE, 2010.

\bibitem[Yan et~al.(2021{\natexlab{a}})Yan, Peng, Fu, Wang, and
  Lu]{yan2021learning}
Bin Yan, Houwen Peng, Jianlong Fu, Dong Wang, and Huchuan Lu.
\newblock Learning spatio-temporal transformer for visual tracking.
\newblock In \emph{Proceedings of the IEEE/CVF international conference on
  computer vision}, pages 10448--10457, 2021{\natexlab{a}}.

\bibitem[Yan et~al.(2021{\natexlab{b}})Yan, Yang, K{\"a}pyl{\"a}, Zheng,
  Leonardis, and K{\"a}m{\"a}r{\"a}inen]{depthtrack}
Song Yan, Jinyu Yang, Jani K{\"a}pyl{\"a}, Feng Zheng, Ale{\v{s}} Leonardis,
  and Joni-Kristian K{\"a}m{\"a}r{\"a}inen.
\newblock Depthtrack: Unveiling the power of {RGBD} tracking.
\newblock In \emph{ICCV}, pages 10725--10733, 2021{\natexlab{b}}.

\bibitem[Yan et~al.(2021{\natexlab{c}})Yan, Yang, K{\"a}pyl{\"a}, Zheng,
  Leonardis, and K{\"a}m{\"a}r{\"a}inen]{yan2021depthtrack}
Song Yan, Jinyu Yang, Jani K{\"a}pyl{\"a}, Feng Zheng, Ale{\v{s}} Leonardis,
  and Joni-Kristian K{\"a}m{\"a}r{\"a}inen.
\newblock Depthtrack: Unveiling the power of rgbd tracking.
\newblock In \emph{Proceedings of the IEEE/CVF International Conference on
  Computer Vision}, pages 10725--10733, 2021{\natexlab{c}}.

\bibitem[Yan et~al.(2024{\natexlab{a}})Yan, Li, Cai, Hao, Jiang, Hu, and
  Xie]{yan2024sanity}
Shilin Yan, Ouxiang Li, Jiayin Cai, Yanbin Hao, Xiaolong Jiang, Yao Hu, and
  Weidi Xie.
\newblock A sanity check for ai-generated image detection.
\newblock \emph{arXiv preprint arXiv:2406.19435}, 2024{\natexlab{a}}.

\bibitem[Yan et~al.(2024{\natexlab{b}})Yan, Xu, Zhang, Hong, Chen, Zhang, and
  Zhang]{yan2024panovos}
Shilin Yan, Xiaohao Xu, Renrui Zhang, Lingyi Hong, Wenchao Chen, Wenqiang
  Zhang, and Wei Zhang.
\newblock Panovos: Bridging non-panoramic and panoramic views with transformer
  for video segmentation.
\newblock In \emph{European Conference on Computer Vision}, pages 346--365.
  Springer, 2024{\natexlab{b}}.

\bibitem[Yan et~al.(2024{\natexlab{c}})Yan, Zhang, Guo, Chen, Zhang, Li, Qiao,
  Dong, He, and Gao]{yan2024referred}
Shilin Yan, Renrui Zhang, Ziyu Guo, Wenchao Chen, Wei Zhang, Hongyang Li, Yu
  Qiao, Hao Dong, Zhongjiang He, and Peng Gao.
\newblock Referred by multi-modality: A unified temporal transformer for video
  object segmentation.
\newblock In \emph{Proceedings of the AAAI Conference on Artificial
  Intelligence}, pages 6449--6457, 2024{\natexlab{c}}.

\bibitem[Yan et~al.(2025)Yan, Han, Tsai, Xue, Fang, Hong, Guo, and
  Zhang]{yan2025crosslmm}
Shilin Yan, Jiaming Han, Joey Tsai, Hongwei Xue, Rongyao Fang, Lingyi Hong,
  Ziyu Guo, and Ray Zhang.
\newblock Crosslmm: Decoupling long video sequences from lmms via dual
  cross-attention mechanisms.
\newblock \emph{arXiv preprint arXiv:2505.17020}, 2025.

\bibitem[Yang et~al.(2022)Yang, Li, Zheng, Leonardis, and Song]{protrack}
Jinyu Yang, Zhe Li, Feng Zheng, Ales Leonardis, and Jingkuan Song.
\newblock Prompting for multi-modal tracking.
\newblock In \emph{ACMMM}, pages 3492--3500, 2022.

\bibitem[Ye et~al.(2022)Ye, Chang, Ma, Shan, and Chen]{ye2022joint}
Botao Ye, Hong Chang, Bingpeng Ma, Shiguang Shan, and Xilin Chen.
\newblock Joint feature learning and relation modeling for tracking: A
  one-stream framework.
\newblock In \emph{European conference on computer vision}, pages 341--357.
  Springer, 2022.

\bibitem[Yu et~al.(2022)Yu, Wang, Vasudevan, Yeung, Seyedhosseini, and
  Wu]{yu2022coca}
Jiahui Yu, Zirui Wang, Vijay Vasudevan, Legg Yeung, Mojtaba Seyedhosseini, and
  Yonghui Wu.
\newblock Coca: Contrastive captioners are image-text foundation models.
\newblock \emph{arXiv preprint arXiv:2205.01917}, 2022.

\bibitem[Zhai et~al.(2022)Zhai, Kolesnikov, Houlsby, and
  Beyer]{zhai2022scaling}
Xiaohua Zhai, Alexander Kolesnikov, Neil Houlsby, and Lucas Beyer.
\newblock Scaling vision transformers.
\newblock In \emph{Proceedings of the IEEE/CVF conference on computer vision
  and pattern recognition}, pages 12104--12113, 2022.

\bibitem[Zhang et~al.(2016)Zhang, Fidler, and Urtasun]{zhang2016instance}
Ziyu Zhang, Sanja Fidler, and Raquel Urtasun.
\newblock Instance-level segmentation for autonomous driving with deep densely
  connected mrfs.
\newblock In \emph{Proceedings of the IEEE Conference on Computer Vision and
  Pattern Recognition}, pages 669--677, 2016.

\bibitem[Zhang et~al.(2020)Zhang, Peng, Fu, Li, and Hu]{zhang2020ocean}
Zhipeng Zhang, Houwen Peng, Jianlong Fu, Bing Li, and Weiming Hu.
\newblock Ocean: Object-aware anchor-free tracking.
\newblock In \emph{Computer Vision--ECCV 2020: 16th European Conference,
  Glasgow, UK, August 23--28, 2020, Proceedings, Part XXI 16}, pages 771--787.
  Springer, 2020.

\bibitem[Zhou et~al.(2023)Zhou, Guo, Hong, Li, Zhang, Ge, and
  Zhang]{zhou2023reading}
Xinyu Zhou, Pinxue Guo, Lingyi Hong, Jinglun Li, Wei Zhang, Weifeng Ge, and
  Wenqiang Zhang.
\newblock Reading relevant feature from global representation memory for visual
  object tracking.
\newblock \emph{Advances in Neural Information Processing Systems},
  36:\penalty0 10814--10827, 2023.

\bibitem[Zhou et~al.(2025)Zhou, Li, Hong, Jiang, Guo, Ge, and
  Zhang]{zhou2025detrack}
Xinyu Zhou, Jinglun Li, Lingyi Hong, Kaixun Jiang, Pinxue Guo, Weifeng Ge, and
  Wenqiang Zhang.
\newblock Detrack: In-model latent denoising learning for visual object
  tracking.
\newblock \emph{arXiv preprint arXiv:2501.02467}, 2025.

\bibitem[Zhu et~al.(2023{\natexlab{a}})Zhu, Lai, Chen, Wang, and
  Lu]{zhu2023visual}
Jiawen Zhu, Simiao Lai, Xin Chen, Dong Wang, and Huchuan Lu.
\newblock Visual prompt multi-modal tracking.
\newblock \emph{arXiv preprint arXiv:2303.10826}, 2023{\natexlab{a}}.

\bibitem[Zhu et~al.(2020)Zhu, Su, Lu, Li, Wang, and Dai]{zhu2020deformable}
Xizhou Zhu, Weijie Su, Lewei Lu, Bin Li, Xiaogang Wang, and Jifeng Dai.
\newblock Deformable detr: Deformable transformers for end-to-end object
  detection.
\newblock \emph{arXiv preprint arXiv:2010.04159}, 2020.

\bibitem[Zhu et~al.(2022)Zhu, Xu, Tang, Wu, Liu, Yang, Wu, and
  Kittler]{zhu2022rgbd1k}
Xue-Feng Zhu, Tianyang Xu, Zhangyong Tang, Zucheng Wu, Haodong Liu, Xiao Yang,
  Xiao-Jun Wu, and Josef Kittler.
\newblock Rgbd1k: A large-scale dataset and benchmark for rgb-d object
  tracking.
\newblock \emph{arXiv preprint arXiv:2208.09787}, 2022.

\bibitem[Zhu et~al.(2023{\natexlab{b}})Zhu, Li, Liu, Wang, Tang, Luo, and
  Huang]{zhu2023tiny}
Yabin Zhu, Chenglong Li, Yao Liu, Xiao Wang, Jin Tang, Bin Luo, and Zhixiang
  Huang.
\newblock Tiny object tracking: A large-scale dataset and a baseline.
\newblock \emph{IEEE transactions on neural networks and learning systems},
  2023{\natexlab{b}}.

\end{thebibliography}
}

\clearpage
\appendix
\section{Appendix}

\begin{table*}[htp]
\begin{center}
    \resizebox{\textwidth}{!}{
        \begin{tabular}{lcccccccccccccccc}
\toprule
\diagbox{Statics}{Datasets}     & LaSOT & GOT-10K & TrackingNet & COCO   & TNL2K & UAVDT & MOT16 & MOT17 & MOT20 & DanceTrack & SportsMOT & TAO   & UVO   & MOSE & OVIS \\
\midrule
Trajectories & 1400  & 10000   & 30600       & 118288 & 1300  & 2593  & 731   & 2388  & 2332  & 419        & 639       & 15997 & 95308 & 3210 & 2482 \\
Videos       & 1400  & 10000   & 30600       & -      & 1300  & 50    & 7     & 21    & 2     & 40         & 45        & 2921  & 6850  & 1307 & 407  \\
Mean Frames  & 2512  & 156     & 472         & -      & 560   & 814   & 759   & 759   & 2333  & 1044       & 635       & 1055  & 89    & 61   & 65   \\
\bottomrule
\end{tabular}

    }
\end{center}
\vspace{-6mm}
\caption{\textbf{Statics of training data.} We combine multiple datasets to create a large scale training data to conduct scaling up experiments.}
\label{tab:appendix_training_data}
\end{table*}

\begin{table}[htp]
\begin{center}
    \resizebox{0.48\textwidth}{!}{
        \begin{tabular}{lccccccccccccccc}
\toprule

& \tabincell{c}{LaSOT\\~\citep{fan2019lasot}} & \tabincell{c}{LaSOT$_{ext}$\\~\citep{fan2019lasot}} & \tabincell{c}{TrackingNet\\~\citep{muller2018trackingnet}} & \tabincell{c}{TNL2K\\~\citep{wang2021towards}} & \tabincell{c}{UAV123\\~\citep{mueller2016benchmark}}  & \makecell[c]{Sum}  \\ 
\midrule
Trajectories & 280 & 150 & 511 & 600 & 123 & 1664 \\
Videos       & 280 & 150 & 511 & 600 & 123 & 1664 \\
Mean Frames  & 2512 & 2395 & 441 & 697 & 1247 & - \\
\bottomrule
\end{tabular}

    }
\end{center}
\vspace{-6mm}
\caption{\textbf{Statics of current benchmarks.} Trajectories in current popular benchmarks are limited.}
\label{tab:appendix_gtrack_curr}
\end{table}

\subsection{GTrack Bench}
\label{sec:x_gtrack}
Existing tracking models~\cite{cui2022mixformer,cui2024mixformerv2,ye2022joint,bai2023artrackv2} tend to evaluate performance on a limited set of benchmarks (about 3-4), as detailed in Table~\ref{tab:appendix_gtrack_curr}. These benchmarks offer limited trajectories and fall short of comprehensively evaluating a model's tracking capabilities. Thus we introduce the GTrack Bench, which consists of 12 challenging benchmarks. Among the 12 benchmarks, 10 are singel object tracking benchmarks, including LaSOT~\cite{fan2019lasot}, LaSOT$_{ext}$~\cite{fan2019lasot}, 
TrackingNet~\cite{muller2018trackingnet},
TNL2K~\cite{wang2021towards},
UAV123~\cite{mueller2016benchmark},
Avist~\cite{noman2022avist},
LaGOT~\cite{mayer2024beyond}, 
LaTOT~\cite{zhu2023tiny}, HOOT~\cite{sahin2023hoot}, and VideoCube~\cite{hu2022global}. LaSOT~\cite{fan2019lasot}, LaSOT$_{ext}$~\cite{fan2019lasot},  TrackingNet~\cite{muller2018trackingnet}, and UAV123~\cite{mueller2016benchmark} are widely used benchmarks for visual object tracking.  TNL2K~\cite{wang2021towards} is a large-scale benchmark for language-guided tracking. Avist~\cite{noman2022avist} focuses on challenging scenes, while LaGOT~\cite{mayer2024beyond} introduces a new benchmark for multi-object tracking. LaTOT~\cite{zhu2023tiny} primarily targets tiny object tracking, and HOOT~\cite{sahin2023hoot} is designed for scenarios with heavy occlusion. VideoCube~\cite{hu2022global} is a large-scale benchmark designed to evaluate models under challenging real-world conditions. Additionally, GTrack Bench includes two datasets from VOS and VIS tasks, MOSE~\cite{ding2023mose} and OVIS~\cite{qi2022occluded}. These datasets emphasize real and complex scenarios, offering more challenging videos. By integrating these datasets, we construct a comprehensive evaluation suite with three times the number of trajectories (4369 in total), allowing for a more thorough assessment of model capabilities in real-world scenarios.

\subsection{Training Data}
\label{sec:x_train}
Currently, state-of-the-art tracking models~\cite{cui2022mixformer,cui2024mixformerv2,ye2022joint,bai2023artrackv2,wei2023autoregressive} are trained on a combination of several datasets, including TrackingNet~\cite{muller2018trackingnet}, LaSOT~\cite{fan2019lasot}, GOT-10K~\cite{huang2019got}, and COCO~\cite{lin2014microsoft}. However, these datasets alone are insufficient for fully training highly capable tracking models. We convert datasets from related tasks into a single object tracking format to create a large-scale training set. These datasets originate from tasks such as single object tracking (LaSOT~\cite{fan2019lasot}, GOT-10K~\cite{huang2019got}, TrackingNet~\cite{muller2018trackingnet}, COCO~\cite{lin2014microsoft}, TNL2K~\cite{wang2021towards}, and UAVDT~\cite{du2018unmanned}), multi-object tracking (MOT16~\cite{milan2016mot16}, MOT17~\cite{dendorfer2021motchallenge}, MOT20~\cite{dendorfer2020mot20}, DanceTrack~\cite{sun2022dancetrack}, SportsMOT~\cite{cui2023sportsmot}), video object segmentation (MOSE~\cite{ding2023mose}), video instance segmentation (OVIS~\cite{qi2022occluded}), and open-world object tracking and segmentation (TAO~\cite{achal2020tao} and UVO~\cite{wang2021unidentified}). Each video in these additional datasets may contain multiple trajectories, as opposed to only one labeled object's trajectory in visual object tracking. Statistics of these datasets are displayed in Table~\ref{tab:appendix_training_data}. By incorporating a substantial number of training trajectories, we expand our dataset to four times its original size, exceeding the capacity of the initial datasets. We conduct our scaling up experiments based on this large scale dataset.

\subsection{Computational Cost}
Our proposed DT-Training and closed-loop scaling strategy do not introduce additional computational overhead during testing. The inference speed of our model is consistent with the baseline models, OSTrack. For instance, our model Ours-B-256-M achieves 93 fps on an NVIDIA 2080 Ti GPU, which is ths same as the baseline OSTrack while delivering superior performance in terms of accuracy. Moreover, Our model maintains strong performance even after compression, highlighting its potential for efficient deployment in real-world scenarios.

\end{document}